\documentclass[manuscript,nonacm]{acmart}
\AtBeginDocument{%
  }

\setcopyright{acmlicensed}
\copyrightyear{2025}
\acmYear{2025}
\acmDOI{XXXXXXX.XXXXXXX}
\acmConference[EAAMO '25]{the fifth ACM Conference on Equity and Access in Algorithms, Mechanisms, and Optimization}{
  2025}{Pittsburgh, USA}
\acmISBN{978-1-4503-XXXX-X/2018/06}

\usepackage{dsfont, amsfonts, bbold, mathrsfs}
\usepackage{mathtools, nccmath}
\usepackage{tikz}
\usepackage{booktabs, multirow}
\usepackage{xcolor}
\usepackage{wrapfig}
\usepackage{url}

\DeclareMathAlphabet{\mathcal}{OMS}{cmsy}{m}{n} 

\newcommand{\DBR}{\mathsf{DBR}}
\newcommand{\DI}{\mathsf{DI}}

\newcommand{\RR}{\mathds{R}}
\newcommand{\EE}{\mathds{E}}

\newcommand{\eps}{\epsilon}
\newcommand{\lam}{\lambda}
\newcommand{\dlow}{\delta_\eps^{\vee}}
\newcommand{\dhigh}{\delta_\eps^{\wedge}}
\newcommand{\de}{\delta_\eps}
\newcommand{\leps}{\lambda_\eps}

\newcommand{\lo}{\hat{f}} 

\newcommand{\XX}{\mathcal{X}}
\newcommand{\Acal}{\mathcal{A}}
\newcommand{\Bcal}{\mathcal{B}}

\newcommand{\PA}{P_\Acal}
\newcommand{\PB}{P_\Bcal}

\newcommand\giv[1][]{\:#1\vert\:}

\providecommand\given{}
\DeclarePairedDelimiterXPP\cE[1]{\EE_P}[]{}{
\renewcommand\given{\nonscript\;\delimsize\vert\nonscript\;}
#1}
\DeclarePairedDelimiterXPP\cP[1]{P}(){}{
\renewcommand\given{\nonscript\;\delimsize\vert\nonscript\;}
#1}
\DeclarePairedDelimiterXPP\uP[1]{P}(){}{
#1}

\makeatletter
\newtheorem*{rep@theorem}{\rep@title}
\newcommand{\newreptheorem}[2]{%
\newenvironment{rep#1}[1]{%
 \def\rep@title{#2 \ref{##1}}%
 \begin{rep@theorem}}%
 {\end{rep@theorem}}}
\makeatother

\newtheorem{proposition}{Proposition}
\newreptheorem{proposition}{Proposition}
\newtheorem{corollary}{Corollary}
\newreptheorem{corollary}{Corollary}
\newtheorem{lemma}{Lemma}
\newreptheorem{lemma}{Lemma}
\newtheorem{example}{Example}

\begin{document}

\title{Reconsidering Fairness Through Unawareness From the Perspective of Model Multiplicity}


\author{Benedikt Höltgen}
\affiliation{%
  \institution{University of Tübingen}
  \city{Tübingen}
  \country{Germany}}
\email{benedikt.hoeltgen@uni-tuebingen.de}

\author{Nuria Oliver}
\affiliation{%
  \institution{ELLIS Alicante Foundation}
  \city{Alicante}
  \country{Spain}}


\begin{abstract}
    Fairness through Unawareness (FtU) describes the idea that discrimination against demographic groups can be avoided by not considering group membership in the decisions or predictions.
    This idea has long been criticized in the machine learning literature as not being sufficient to ensure fairness.
    In addition, the use of additional features is typically thought to increase the accuracy of the predictions for all groups, so that FtU is sometimes thought to be detrimental to all groups.
    In this paper, we show both theoretically and empirically that FtU can reduce algorithmic discrimination without necessarily reducing accuracy. We connect this insight with the literature on Model Multiplicity, to which we contribute with novel theoretical and empirical results. Furthermore, we illustrate how, in a real-life application, FtU can contribute to the deployment of more equitable policies without losing efficacy. Our findings suggest that FtU is worth considering in practical applications, particularly in high-risk scenarios, and that the use of protected attributes such as gender in predictive models should be accompanied by a clear and well-founded justification.
\end{abstract}




\maketitle


\section{Introduction}

Nurtured by the ever-increasing availability of data and compute, as well as a widespread fondness of quantification in general \cite{porter1995} and Big Data in particular \cite{boyd2012}, machine learning-based systems are increasingly used to assist decision making in a variety of areas of society, including in high-stakes domains such as healthcare, finance, law enforcement, and the provision of social services. 
Leveraging data is often seen not only as a way to improve decisions but also as a way to avoid human bias and enhance objectivity and fairness. 
However, there is mounting evidence of a variety of problems that can arise in different contexts as a result of this trend \cite{wang2022}.
One such problem is unintended algorithmic discrimination, which can arise for different reasons, including biased training data, flawed model assumptions, unequal representation of different demographic groups, or misuse of the models. As a result, algorithmic decisions may systematically disadvantage certain individuals or communities grouped by their protected attributes, such as gender, race, age, or religion. These concerns have prompted a substantial body of research focused on understanding, detecting, and mitigating algorithmic biases \cite{fairmlbook}.
A seemingly straightforward approach to preventing algorithmic discrimination across groups is the exclusion of protected attributes from the set of features that the model can ``see". This idea is known as \emph{Fairness through Unawareness} or FtU \cite{fairmlbook}.
However, the algorithmic fairness literature has shown that avoiding the use of protected attributes in the models is generally not sufficient to prevent discrimination \cite{pedreshi2008, calders2010}: other variables can encode group membership information and thus their use could lead to discrimination.
Furthermore, FtU is often criticized not only for being an unreliable approach to achieve fairness, but also for potentially reducing the predictive performance of the model for all groups \cite{corbett2023}.\footnote{It has been noted that such a reduction in accuracy need not lead to a reduction in utility \cite{coots2023}.}

Such results align with a common intuition in the Big Data era: that more data (and attributes) lead to better, fairer, and more objective decisions \cite{boyd2012}.
A recent practical example of this way of thinking can be found in Austria, where gender was used as an input in an algorithm designed to allocate job training programs to unemployed individuals.
Officials dismissed criticisms regarding potential discriminatory outcomes by arguing that the algorithm's predictions reflect the actual chances individuals face in the labour market (as cited in \citet{allhutter2020}).
More generally, many arguments in favour of including protected attributes rely on the idea that machine learning models aim to estimate ``true" probabilities from data, which is a questionable assumption \cite{holtgen2024position}.
As a result of this assumption, modelling choices may be accepted with limited scrutiny or justification, despite their significant implications \cite{moss2022}.
However, there is growing evidence that policy outcomes are highly sensitive to modelling decisions, including the choice of thresholds, input variables, and specific design parameters \cite{bach2023,kern2024}.
In parallel, emerging research on \emph{model multiplicity} highlights that different models trained on the same data can perform similarly in terms of accuracy, yet differ substantially in their behaviour, particularly with respect to fairness \cite{black2022}.
This perspective highlights that there is no clear hierarchy between models in terms of accuracy and that modelling choices need to be checked against alternative choices.

In light of these concerns, we revisit the commonly held scepticism toward \emph{Fairness through Unawareness} (FtU). While we do not suggest that FtU offers a simple solution to achieve algorithmic fairness, we demonstrate that, contrary to prevailing assumptions, omitting protected attributes does not necessarily harm predictive performance and can contribute to more equitable outcomes. 
To substantiate these claims, we analyse the relationship between data, model class, and fairness, both from a theoretical and from an empirical perspective. 
While we derive general insights, we place particular emphasis on logistic regression because of its broad and extensive use in practice to build models from tabular data. Regarding fairness, we focus on disparate impact (DI) or statistical/demographic parity, which compares the positive classification rates for different groups according to their protected attribute. 
Our findings suggest that including protected attributes can be especially problematic in the context of logistic regression, as their use by such models can easily exacerbate disparities between protected groups. 
By linking the concepts of FtU and model multiplicity, we argue for a shift in perspective: fairness should not be seen as a constraint or a trade-off, but as a deliberate modelling choice that can lead to more equitable outcomes without necessarily sacrificing predictive performance.

\begin{figure}
    \centering
    \label{fig:fig1}
\end{figure}

In sum, the main \textbf{contributions} of this work are as follows:
\begin{itemize}
    \item We analyse the relationship between disparate impact and base rate difference in the data both for any classifier and for logistic regression. In the case of logistic regression, we demonstrate the risk of exacerbated unfairness depending on the classification thresholds. 
    \item We contribute to the understanding of model multiplicity and algorithmic fairness by providing mathematical results that characterise the potential for multiplicity and connect it to fairness: (1) general upper bounds on model multiplicity and changes in disparate impact which are tight for unconstrained models; and (2) lower bounds on achievable reductions in disparate impact based on FtU for logistic regression. 
    \item We empirically demonstrate how unaware models can yield similar levels of accuracy than aware models with significant improvements in fairness in multiple settings.
    \item We highlight the practical implications on our work by discussing a real-world example related to employment programs in a European country. 
\end{itemize}

The \textbf{structure} of this article is as follows.
In Section~\ref{s:background}, we provide an overview of related work on fairness, less discriminatory alternatives, and model multiplicity.
In Section~\ref{s:calibrated}, we highlight the importance of modelling choices by demonstrating that there is no straightforward connection between disparate impact and differences in base rates, even assuming well-calibrated models. We also show that disparate impact can often be higher than the base rate difference for \emph{aware} models (\emph{i.e.} models using protected attributes), especially for models based on logistic regression.
In Section~\ref{s:ldas}, we frame our work from the perspective of model multiplicity. We provide new theoretical upper bounds and relate them to disparate impact and to fairness through unawareness. In the case of logistic regression, we also provide theoretical lower bounds for reducing disparate impact. We illustrate our insights empirically in Section~\ref{s:empirical}, showing that FtU can provide fairer models with almost equal accuracy across multiple datasets, protected attributes, and model classes. 
Section~\ref{s:ams} illustrates the impact of our work in a real-world scenario where an aware logistic regression model was used to assign unemployed individuals to job training programs. We show how in this case an unaware model would be equally performative but fairer. Finally, the main conclusions and lines of future work are described in Section~\ref{s:conclusion}.


\section{Related work}
\label{s:background}



\subsection{Algorithmic fairness and disparate impact}

Under the influence of legal, economic, and philosophical notions of discrimination as well as several high-profile investigations into existing software, an extensive body of work on algorithmic fairness has introduced a variety of metrics to quantify and address discrimination in machine learning systems \cite{fairmlbook}.
One line of work focuses on error rate disparities, \emph{i.e.}, the differences in false positive or false negative rates across demographic groups. However, some scholars have argued that error rates should serve as \emph{diagnostic tools} rather than direct targets for fairness interventions \cite{fairmlbook}. 
Another influential line of work draws on legal and policy traditions, particularly anti-discrimination law in the United States, introducing the concept of \emph{disparate impact}---also referred to in the machine learning literature as \textit{demographic or statistical parity} \cite{barocas2016, feldman2015}. Disparate Impact focuses on algorithmic outcomes rather than errors: it compares the rates at which individuals from different demographic groups receive positive classifications (\emph{e.g.}, job offers, university admissions, loan approvals). If one group consistently receives favourable outcomes at a lower rate than other groups, this may be evidence of systemic bias.\footnote{These outcome disparities are sometimes quantified by means of the \emph{disparate impact ratio} and evaluated using heuristics like the ``80\% rule"---which, however, is only a ``crude test" \cite{raghavan2024}.} 
%
While disparate impact has the advantage of aligning closely with legal standards and public perceptions, it also faces criticism, particularly in machine learning contexts. 
Enforcing strict parity in classification outcomes may lead to a reduction in accuracy and thus overall ``utility'' \cite{hardt2016}, especially when base rates differ across groups. This trade-off can result in decisions that, while statistically balanced, may harm individuals or reduce efficiency without delivering clear benefits \cite{corbett2023}. Moreover, socio-technical critiques point out that measures such as disparate impact that focus on classification rates, just like error-focused measures, only consider model properties and may fail to result in just outcomes when actually deployed \cite{selbst2019,kuppler2022}. 
%
Nonetheless, disparate impact has a more straightforward connection to the real-world consequences of models than error rates. In this work, we assume that reducing disparate impact is often desirable as long as it does not imply a notable reduction in accuracy, broadly in line with calls for non-ideal approaches to fairness \cite{fazelpour2020} and comparative (rather than transcendental) approaches to justice \cite{sen2010}. 
Indeed, this perspective reflects the original motivation behind the notion of disparate impact; in contrast, its criticism in algorithmic fairness often concerns the full elimination of disparate impact, which reflects its common treatment as a mathematical formula that is exchangeable with other metrics.




\subsection{Less discriminatory alternatives and model multiplicity}

In legal doctrine under U.S. anti-discrimination law, the concept of disparate impact is closely related to the idea of a \emph{Less Discriminatory Alternative}: a policy or decision-making process that leads to disparate impact can be justifiable as long as there exists no alternative that meets the same legitimate objectives while producing less discriminatory outcomes. This principle plays a central role in Title VII of the Civil Rights Act and has informed a wide range of employment and civil rights cases \cite{barocas2016}. 
In the context of machine learning, less discriminatory alternatives can be operationalized as a fairness-aware criterion for model selection: if two models perform similarly regarding predictive accuracy but one model has lower disparate impact than the other, the fairer model could be considered as the less discriminatory alternative and thus would be both ethically preferable and potentially legally required in regulated environments \cite{barocas2016}.

Although central to disparate impact, the concept of less discriminatory alternatives has only recently gained attention in the ML literature, particularly in the context of \emph{model multiplicity} (MM) \cite{black2024short, gillis2024}.
MM describes the phenomenon where multiple different models can achieve comparable accuracy on the same task.
While the underlying idea---sometimes called \emph{Rashomon effect}---has been known for decades \cite{breiman2001}, only in recent years has it become a focus of systematic investigation \cite{marx2020}.
Despite promising recent work \cite{semenova2023}, this area of research is still in its early stages.

An intriguing aspect of MM is that models with similar accuracy can exhibit very different behaviours regarding robustness, fairness, or interpretability \cite{coston2021,black2022,rudin2024}. 
By virtue of MM, it is often possible to select fairer models without losing accuracy, \emph{i.e.} less discriminatory \emph{algorithms} (LDAs) \cite{black2024}.
In such cases, the existence of accurate and fairer alternatives underscores the need to examine the modelling pipeline beyond aggregate performance metrics. From a fairness perspective, MM implies that selecting one model over another is not only a neutral performance-based decision, as it can have a differential real-world impact across demographic groups. 
%
In this paper, we investigate how Fairness through Unawareness (FtU) can, in some cases, provide a less discriminatory alternative without requiring explicit fairness optimisation, which has been investigated in other works \cite{coston2021,gillis2024}. We examine whether models that exclude protected attributes from their input can still maintain competitive predictive performance while reducing disparate impact, particularly in the context of logistic regression, a widely used model in real-world scenarios. 



\section{Calibrated aware models can exacerbate inequality}
\label{s:calibrated}

\begin{quote}
    \emph{``Predictive models trained with supervised learning methods are often good at calibration [...] But calibration also means that by default, we should expect our models to faithfully reflect disparities found in the input data."} 
    \cite[p. 21]{fairmlbook}
\end{quote}

Reflected in this quote from the most prominent textbook on
algorithmic fairness is the frequent assumption that disparities present in the data will be carried over into calibrated machine learning models by default. 
While this assumption may hold true for probabilistic models, it does not apply to classifiers derived from these models. 
In this section, we demonstrate that, perhaps counterintuitively, decisions based on calibrated probabilities can actually worsen the inequalities reflected in the base rates of labels. 
We also show that this effect is especially pronounced in policies relying on logistic regression models.

We assume an input space $\XX$ and a, possibly empirical, joint probability distribution $P$ over random variables $(X,Y)$ which can take binary values in $\XX \times \{0,1\}$.
Take a predictor $f : \XX \to [0,1]$
and a corresponding classifier 
\begin{equation}\label{eq:thresh_class}
    F : \XX \to \{0,1\},\ x \mapsto \mathbb{1}[f(x) \geq 0.5].
\end{equation}
Assume two groups that form a partition $\{\Acal, \Bcal\}$ of $\XX$.\footnote{This is the case when the input features contain a protected attribute encoding group membership, i.e., for group-aware models.}
The purpose of this section is to analyse how the disparate impact (DI)
\begin{equation}\label{eq:disp_imp}
    \DI(F) := 
    \cE*{F(X) \given\ X \in \Acal} - \cE*{F(X) \given\ X \in \Bcal}
\end{equation}
relates to the difference in base rates (DBR)
\begin{equation}\label{eq:br_diff}
    \DBR := 
    \cE*{Y \given\ X \in \Acal} - \cE*{Y \given\ X \in \Bcal}.
\end{equation}
In this paper, we assume that $\Bcal$ is the disadvantaged group, \emph{i.e.}, $DBR \geq 0$.

\subsection{General case}

The first observation is that for calibrated predictors, the label rates in the data are reflected in the means of the predictors.
In the standard sense of \citet{chouldechova2017}, a predictor $f$ is \textit{calibrated} on both groups if 
\begin{equation}\label{eq:calib_groups}
    \forall v \in [0,1], \ast \in \{\Acal, \Bcal\}:\quad
    \cE*{Y \given f(X)=v, X \in \ast} = v.
\end{equation}
Now, if $f$ is calibrated in this sense, then the average prediction of $f$ on either group trivially equals the average outcome for that group (simply by integrating over $v$):
\begin{equation}\label{eq:calib_groups_global}
    \forall \ast \in \{\Acal, \Bcal\}:\quad
    \EE_P \left[ f(X) \giv X \in \ast \right] = \EE_P \left[Y \giv X \in \ast\right].
\end{equation}

While all predictors that are calibrated on a given dataset will have the same mean, equal to the base rate, these different predictors can lead to different classification rates.
We illustrate this with a simple toy example.
\begin{example}\label{ex:calib_pred}
    Take $\XX = \{x_1, x_2, x_3, x_4\}$ with $P$ defined via $\forall x \in \XX : P(x) = 0.25$ and $P(Y=1 \giv X=x) = 0.4, 0.4, 0.55, 0.7$ for $x = x_1, x_2, x_3, x_4$, respectively.
    Then the predictors $f_1, f_2, f_3$ as defined in Table~\ref{tab:calib_pred_example}a are all calibrated in the standard sense (without groups) while leading to very different classifications (Table~\ref{tab:calib_pred_example}b).
\end{example}

\begin{table}[h]
    \centering
    \caption{
    Left (table a): Calibrated predictors for Example~\ref{ex:calib_pred}, where $p(x)=P(Y=1 \giv X=x)$ is the true probability.
    $f_1$ is given by joining $x_3$ with $x_1$ and $x_2$; $f_2$ given by joining $x_1$ with $x_3$ and joining $x_2$ with $x_4$; $f_3$ is given by the constant base rate predictor. All of these predictors are calibrated.
    Right (table b): Classifiers derived by the models in table a, where $B$ is the Bayes optimal classifier.}
    \begin{tabular}{c|c|c|c|c}
        a & $x_1$ & $x_2$ & $x_3$ & $x_4$ \\
        \hline
        $p(x)$ & 0.4 & 0.4 & 0.55 & 0.7 \\
        \hline
        $f_1(x)$ & 0.45 & 0.45 & 0.45 & 0.7 \\
        \hline
        $f_2(x)$ & 0.475 & 0.55 & 0.475 & 0.55 \\
        \hline
        $f_3(x)$ & 0.5125 & 0.5125 & 0.5125 & 0.5125 \\
        \hline
    \end{tabular}
    \hspace{1cm}
    \begin{tabular}{c|c|c|c|c}
        b & $x_1$ & $x_2$ & $x_3$ & $x_4$ \\
        \hline
        $B(x)$ & 0 & 0 & 1 & 1 \\
        \hline
        $F_1(x)$ & 0 & 0 & 0 & 1 \\
        \hline
        $F_2(x)$ & 0 & 1 & 0 & 1 \\
        \hline
        $F_3(x)$ & 1 & 1 & 1 & 1 \\
        \hline
    \end{tabular}
    \label{tab:calib_pred_example}
\end{table}

We turn next to the quantities of interest: disparate impact and difference in base rates.
First, we consider the most general case, with arbitrary $f$, $\alpha$, and $P$.
It is important to note that disparate impact depends on the ratio of predictions above/below the decision threshold, whereas the base rate difference depends on the means of the predictions (assuming calibration).
We can formalise this in terms of the distributions of the predictions for each group, $\PA$ and $\PB$, where
\begin{equation}
    P_\ast(v) := P(f(X)=v \giv X  \in \ast) \quad \text{for} \quad v \in [0,1],\ \ast \in \{\Acal, \Bcal\}.
\end{equation}
Then, for random variables $V_a, V_b \sim \PA, \PB$ describing the group-wise predictions, the disparate impact is given by
\begin{equation}\label{eq:disp_imp2}
    \DI(F) 
    = P(F(X) = 1 \giv X \in \Acal) - P(F(X) = 1 \giv X \in \Bcal)
    = \PA(\{V_a \geq 0.5\}) - \PB(\{V_b \geq 0.5\})
\end{equation}
and the difference in base rates by
\begin{equation}\label{eq:br_diff2}
    \DBR = 
    \EE_{\PA}[V_a] - \EE_{\PB}[V_b].
\end{equation}
The difference in base rates can intuitively be understood as the balance of how much and how far the probability mass is moved to the left vs the right of the decision boundary when moving from $\PB$ to $\PA$.\footnote{This means that if $\PA$ `dominates' $\PB$ in the sense of $\exists T:[0,1] \to [0,1]$ with $T(v) \geq v \ \forall v \in [0,1]$ s.t. $T(V_b) \sim \PA$, then the base rate difference equals the earth mover (Wasserstein-1) distance.}
The disparate impact, on the other hand, only depends on the balance of how much probability mass passes over the decision threshold.

The base rate difference and the disparate impact can be directly related under specific conditions, for example, when $\PA$ is higher on one side of the threshold and $\PB$ is higher on the other (proofs can be found in Appendix~\ref{app:proofs}):
\begin{proposition}[Relation between Base Rates and Disparate Impact]
    \label{prop:brd_di}
    Assume $f$ is calibrated as in (\ref{eq:calib_groups}), satisfying $\forall v \in [0,0.5): \PB(v) \geq \PA(v)$ and $\forall v \in [0.5,1] : \PA(v) \geq \PB(v)$.
    Then for $F$ defined as in (\ref{eq:thresh_class}), $\DI(F) \geq \DBR$.
\end{proposition}

\subsection{Special case: Logistic Regression}
\label{ss:aware_LR}

We now investigate the relationship between disparate impact and the difference in base rates in the case of logistic regression.
Logistic regression is a widely used type of statistical method for modelling the probability of a binary outcome based on one or more predictor variables. Its popularity is due to its simplicity, interpretability,  and effectiveness in tackling classification problems. 
The model class consists of linear models with a sigmoid output $\sigma$, 
\begin{equation}
    f: \RR^n \to [0,1],\ (x_1, ..., x_n) \mapsto \sigma \left( c_0 + \sum_i c_i x_i\right)
\end{equation}
with $\XX \subset \RR^n$ and the sigmoid function defined by $\sigma: z \mapsto \frac{1}{1+e^{-z}}$.

Training a logistic regression model consists of finding the values of the model's coefficients $c_0, ..., c_n \in \RR$ that best fit the data by minimising a logistic loss function (equivalent to maximising the likelihood) using optimisation techniques like gradient descent. If a protected attribute is part of the input space, there will be a corresponding coefficient $c_i \in \RR$ which determines how the protected attribute affects the (pre-sigmoid) logit function, $\lo(x) := c_0 + \sum_i c_i x_i$.
Therefore, for binary protected attributes where $x_i$ can be 0 or 1, being part of the disadvantaged group ($x_G$ = 1) implies adding a term $c_G$ to the logit without further interaction with the rest of the variables.

\begin{wrapfigure}{r}{0.34\linewidth}
    \centering
    \vspace{-1em} 
    \includegraphics[width=\linewidth]{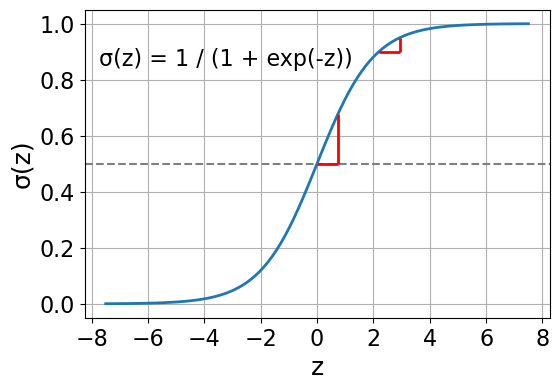}
    \caption{Plot of the sigmoid function used in logistic regression. Note that the slope is steepest around $z=0$ and satisfies $\sigma(0)=0.5$.}
    \label{fig:sigmoid_plot}
    \vspace{-1em} 
\end{wrapfigure}
We focus on the case where $P(Y \giv X, G=\ast)$ differs significantly between groups, as this makes the existence of a fixed global coefficient $c_G$ particularly problematic.\footnote{In the case where $P(Y \giv X, G=\ast)$ is the same across groups but $P(X \giv G=\ast)$ is different, the choice of model class may not have as immediate an effect.
However, this choice still affects the resulting predictive distribution, which has an effect on the disparate impact as shown above.}
A difference in base rates means that, if the model is calibrated for both groups (\ref{eq:calib_groups_global}), the coefficient of the variable that corresponds to the group attribute will penalise all members of the disadvantaged group because the coefficient for that group will reduce the logit by a fixed value $c_G$, before applying the sigmoid, which can then contribute to an even larger disparate impact. 
To understand why, consider the shape of the sigmoid function displayed in Figure~\ref{fig:sigmoid_plot}: 
The curve is steeper around the classification threshold at $\sigma(z)$ = $0.5$ than for other values of $z$, as highlighted by the red lines. Therefore, small changes in $z$ near the classification threshold would lead to significant changes in the value of the sigmoid. 

Consider the following example with a logistic regression classifier: 
Two individuals have feature vectors $x^b$ and $x^a$ that are identical except for their group membership, such that person $b$ belongs to group $\Bcal$ and person $a$ belongs to group $\Acal$.
Therefore, the only element that differs between the two individuals' predictions is the coefficient corresponding to the variable that denotes which group they belong to. Changes in this coefficient will shift the logit value $z$. 
Imagine the output of the logistic regression function (predicted probabilities) for each of them is $\sigma(z^b)$ = $0.95$ and $\sigma(z^a)$ = $0.9$. To obtain the logit values $z^a$ and $z^b$, we need to apply the inverse sigmoid $z = \sigma^{-1}(p) = ln(\frac{p}{1-p})$. In the example, $\sigma^{-1}(0.95) \approx 2.94$ and $\sigma^{-1}(0.90) \approx 2.20$, such that their difference is  $\approx 0.75$. This difference comes from the assumed difference in base rates between groups in this example. 
This suggests that group $\Acal$ has a group offset that raises the logit by 0.75 when compared to group $\Bcal$ for the same input features. Now, let's consider an individual in group $\Bcal$ that has a predicted probability right at the classification threshold at $\sigma(z) = 0.5$ (therefore, logit $z=0$); and another individual with exactly the same feature vector but belonging to group $\Acal$. The logit would increase by 0.75 because of the group effect, such that $z_a = z_b + 0.75 = 0 + 0.75$, $\sigma(0.75) \approx 0.68$. 
This means that the same features lead to higher predictions for members of $\Acal$ due to the group offset: for every individual in group $\Acal$ with scores between $0.5$ and $0.68$, there is a counterpart in $\Bcal$ who would fall below the threshold, even though their other features are identical. 

Thus, a global group coefficient that is meant to account for overall base rate differences between groups can lead to large local effects, especially in the vicinity of the decision threshold, due to the non-linear shape of the sigmoid function. In the example, even a moderate difference of 0.05 closer to the margins corresponds to a large difference of 0.18 near the threshold, which can lead to a large disparate impact. 


\begin{figure}[h]
    \centering
    \includegraphics[width=.49\linewidth]{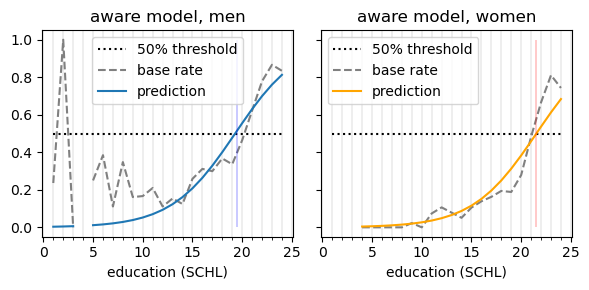}
    \includegraphics[width=.49\linewidth]{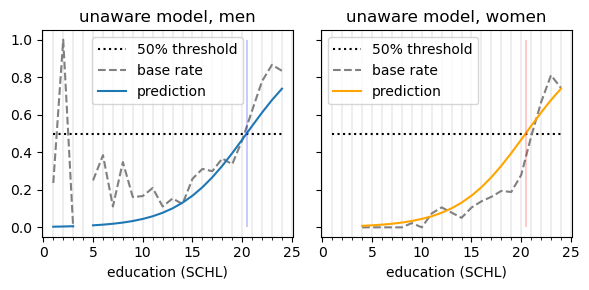}
    \caption{Predictions for aware and unaware logistic regression models on the ACS Income NY dataset. The aware model uses education and sex as input variables. The unaware model only considers education, measured in years. Colored vertical lines correspond to the 50\% cut-off. Note how the aware logistic regression model requires much higher education values for women (yellow/red) than for men (blue) to be selected. This phenomenon is not observed in the unaware model.}
    \label{fig:1-feat}
\end{figure}

This effect can be further illustrated with an example based on real data, namely the US Census data from New York, obtained from the ACS Income dataset from the \texttt{folktables} package \cite{ding2021} (see Section~\ref{ss:empirical_setup}).
In this case, the target variable is whether a person has an annual income of at least 50k.
For illustration purposes, we restrict the logistic regression model to include only two variables: the most predictive attribute (education, in number of years) and the protected binary attribute (sex). 
Two logistic regression models using default hyperparameters are trained: one \emph{aware} model, including education and sex, and one \emph{unaware} model that uses education as the only input variable. 
The results are depicted in Figure~\ref{fig:1-feat}:
The aware model ``pushes" the predictions for men and women apart, meaning that men need an education score of 20 (equivalent to an associate's degree), whereas women need an education score of 22 (equivalent to a Master's degree) to obtain a positive classification.
Income for men is over-predicted around the threshold (the blue line being above the grey dashed base rate line), whereas for women it is under-predicted (the yellow line being below the grey dashed base rate line).
Conversely, in the case of the unaware model, which does not include sex as an input variable, the blue line on the left and the orange line on the right are necessarily the same.
In this case, men are predicted correctly at the threshold while women are slightly over-predicted (the yellow line being above the grey dashed base rate line).
Overall, this results in a similar accuracy for the unaware model (69.9\%), compared to the aware model (70.0\%), while significantly reducing the disparate impact (more details can be found in Appendix~\ref{app:single-feat}).\footnote{While the single feature case is an unrealistic illustration, the fact that the accuracy is not much lower than when using all features indicates that the feature drives most of the prediction. This already suggests that similar results occur in full datasets; we turn to this in Section~\ref{s:empirical}.}

The phenomenon that multiple models can have similar accuracy while differing in other properties, such as fairness, falls under the notion of model multiplicity.
Before empirically investigating larger datasets, we shed light on this effect from the perspective of model multiplicity for a better theoretical understanding.


\section{Less Discriminatory Algorithms through Unawareness}
\label{s:ldas}
In this section, we show that large differences between models with similar accuracy can be explained theoretically and that datapoints with predictions close to the decision threshold is particularly important for this.
We provide the first tight bounds on model multiplicity and, based on that, derive bounds on the difference in disparate impact.
We first prove general upper bounds on how much models with a given level of accuracy can differ, followed by a lower bound specifically for logistic regression classifiers.


\subsection{General upper bounds on model multiplicity}
\label{ss:upper_bounds}

We present new upper bounds on the extent to which models with the same accuracy can diverge. We show that this divergence---model multiplicity---can be especially pronounced when label ratios for certain inputs are close to 50\%.
We begin by establishing general results on model multiplicity and then explore their implications for disparate impact and unaware models that do not use the protected attribute.

We assume an empirical distribution $P$ over $X,Y$ describing the available data; based on this we define the conditional $p(x) : x \mapsto P(Y=1 \giv X=x)$ and marginal measure $\mu(U) : 2^\XX \to [0,1], U \mapsto \int_U P(x) dx$. 
Furthermore, let $B : \XX \to \{0,1\}$ be a Bayes-optimal classifier (with $B(x) = 0$ for $p(x)<0.5$ and $B(x) = 1$ for $p(x)>0.5$).
Now in line with the closest previous work \cite{black2022}, we define $L$ as the 0-1-loss
\begin{equation}
    L(y_1, y_2) := \begin{cases}
        0 & \text{ if } y_1 = y_2 \\
        1 & \text{ otherwise}.
    \end{cases}
\end{equation}
We define the accuracy of a model $F$ as
\begin{equation}
    Acc(F) := 1 - \EE_P[L(F(X), Y)]
\end{equation}
and the disagreement between two models $F,G$ as 
\begin{equation}
    d(F,G) := \mu(\{x \in X : F(x) \neq G(x)\}).
\end{equation}
For $\mathcal{F}_\XX := 2^\XX$ the set of binary classifiers on $\XX$, let 
\begin{equation}
    R_\eps(F) := \{G \in \mathcal{F}_\XX : Acc(G) \geq Acc(F) - \eps\}
\end{equation}
be the $\eps$-Rashomon set of $F$.
For $\lam \in [0,0.5]$, let 
\begin{equation}
    U(\lambda) := \{x \in \XX: |p(x) - 0.5| = \lambda\}
    \quad \text{and} \quad
    m(\lam) := \mu(U(\lam))
\end{equation}
measure how much each distance from $0.5$ occurs, which is non-zero only on the set of occurring distances $\Lambda := \{|p(x) - 0.5| : x \in \XX\}$.
Then we define new quantities 
\begin{equation}
    e(\lam) := \sum_{\lam' < \lam} 2 \lam' \cdot m(\lam')
    \quad\text{and}\quad
    \leps := \arg \max_{\lam \in \Lambda} \{ \lam : e(\lam) \leq \eps\},
\end{equation}
which are illustrated in Figure~\ref{fig:lambda_eps} in Appendix~\ref{app:proofs}.
Intuitively, $e(\lambda)$ is the additional error if for all points with label ratios closer to $50\%$ than $\lambda$, the suboptimal classification is taken, and $\leps$ is the largest $\lambda$ such that this error does not exceed $\eps$.
With these definitions at hand, we can prove a tight upper bound on model multiplicity, which is proved through a lemma on randomised classifiers.
The dependence of our tight bound on $\leps$ and $e(\leps)$ demonstrates how important input space regions with a near-$50\%$ label ratio are for model multiplicity.

\begin{proposition}[Upper Bound on Multiplicity]
    \label{prop:max_rash}
    \begin{equation}\label{eq:max_rash}
        \max_{F \in R_\eps(B)} d(F,B)
        \leq \frac{\eps - e(\leps)}{2 \leps} + \sum_{\lam < \leps} m(\lam).
    \end{equation}
    This bound is the tightest possible bound with only access to the cumulative distribution of $p(x)$.
\end{proposition}

\begin{lemma}
    \label{lem:rand_class}
    The bound (\ref{eq:max_rash}) can be achieved in any setting for every $\eps$ by a \emph{randomised} classifier defined as
    \begin{equation}
        F_\eps(x) = \begin{cases}
            1-B(x) & \text{ for } |p(x)-0.5| < \leps \\
            1-B(x) \text{ with probability } \frac{\eps - e(\leps)}{2 \leps \cdot m(\leps)} & \text{ for } |p(x)-0.5| = \leps \\
            B(x) & \text{ for } |p(x)-0.5| > \leps.
        \end{cases}
    \end{equation} 
\end{lemma}

This means the maximum is achieved if the predictions with $p(x)$ closest to $0.5$ are flipped until the accuracy decreases by $\eps$.
All proofs, as well as illustrations of these results and the quantity $e(\lambda)$ can be found in Appendix~\ref{app:proofs}.


We now relate this bound to a \textbf{previous result}: \citet{black2022} provide a formal upper bound on how much a model can disagree with another model when the difference in accuracy is controlled. We can also recover their result in the proof of their Theorem A.2, that 
\begin{equation}\label{eq:black_bound}
    d(F,B) \leq \frac{Acc(B)-Acc(F)}{2c}
\end{equation}
under the assumption\footnote{Black et al. assume $|p(x) - 0.5| > c$ but it is not necessary to assume that the inequality is strict.}  
\begin{equation}\label{eq:black_ass}
    |p(x) - 0.5| \geq c > 0\ \forall x \in \XX.
\end{equation}

\begin{corollary}
    \label{cor:black}
    Under assumption (\ref{eq:black_ass}),
    \begin{equation}\label{eq:black_cor}
    \max_{F \in R_\eps(B)} d(F,B) \leq \frac{\eps}{2c}.
    \end{equation}
\end{corollary}



While the above bounds involved an optimal classifier, we can also give bounds for \textbf{two non-optimal models}.

\begin{proposition}[Multiplicity Bound for Non-optimal Models]\ \\
\label{prop:two_class}
    We can bound the disagreement between two classifiers $F \in R_\eps(B)$ and $G \in R_\delta(B)$ 
    via
    \begin{equation}\label{eq:max_rash_nonopt}
    d(F,G) \leq \max_{H \in R_{\eps + \delta}(B)} d(H,B).
    \end{equation}
\end{proposition}

That is, the maximum possible disagreement between $F$ and $G$ is no greater than the largest disagreement that any model $H$ within accuracy drop $\eps + \delta$ could have with $B$ itself. 
We now connect this bound to \textbf{disparate impact}.
\begin{proposition}[Upper Bound on Disparate Impact]
\label{prop:disp_imp}
    For two classifiers $F$ and $G$, the difference in their disparate impact can be bounded based on their disagreement and the size of the smaller group:
    \begin{equation}\label{eq:di_diff}
    | \DI(F) - \DI(G) | \leq \frac{d(F,G)}{\min_{\ast \in \{\Acal,\Bcal\}} \uP{X \in \ast}}.
    \end{equation}
\end{proposition}

That is, the potential difference in disparate impact between two models increases with their disagreement, and it is inversely proportional to the size of the smallest protected group.  
We can also specify a bound for the context of \textbf{Fairness through Unawareness}. The above bound is not tight when an unaware model is involved, unless for all $\bigcup_{\lam < \leps} U(\lam)$, the two groups are separable in $U(\lam)$ by the protected attribute alone.
If this is not the case, then the unaware model will give the same decision for some $x_1, x_2 \in \bigcup_{\lam < \leps} U(\lam)$ from different groups; so flipping the decisions on both $x_1$ and $x_2$, which is necessary for equality in (\ref{eq:max_rash}), will not lead to a proportional rise in DI, \emph{i.e.}, there is no equality in (\ref{eq:di_diff}).
However, we can show that half the bound is achievable for some $\eps$:
\begin{proposition}[Achievable DI for Unaware Models]
    \label{prop:unaware}
    Fix $P$, an unaware model $F$, and some $\eps > 0$.
    Then there is a model $G \in R_\eps(F)$ with 
    \begin{equation}\label{eq:di_diff_lower}
        | \DI(F) - \DI(G) | \geq \frac{1}{2 \cdot \max_{\ast \in \{\Acal,\Bcal\}} \uP{X \in \ast}} \max_{H \in R_{\eps + \delta}(B)} d(H,B).
    \end{equation}
\end{proposition}





\subsection{A lower bound for logistic regression}
\label{ss:lower_bound}

We now turn to the specific case of LR models.
The following result provides a lower bound on the reducible DI for a fixed change in accuracy. This bound depends on the coefficient of the protected attribute and the calibration plots for both groups, including bin sizes. Similar to the results presented above, it crucially depends on changes in classification decisions near the decision threshold, highlighting the importance of data points near the decision threshold. 

\begin{proposition}[Lower Bound on DI for LR]
    \label{prop:unaware_LR}
    Let $F : \XX \to \{0,1\}$ be an aware classifier based on an LR model $f$ with coefficient $c_{G} < 0$ for the PA, i.e., the disadvantaged group gets a pre-sigmoid penalty of $c_{G}$.
    Further, let 
    \begin{equation}
        Q(c_G) := \left\{x \in \Bcal : f(x) \in [\sigma(c_{G}), 0.5]\right\}
    \end{equation}
    denote the set of points that are in Group $b$ \emph{and} get predictions between $\sigma(c_{G})$ and $0.5$ by $f$.
    Then the corresponding unaware classifier $F'$ based on setting $c_{PA} \leftarrow 0$ in $F$ has accuracy bounded by
    \begin{equation}\label{eq:unaware_LR_acc}
        | Acc(F) - Acc(F') | \leq 2 \sigma(-c_{G}) \cdot P(X \in Q(c_G))
    \end{equation}
    and disparate impact given by
    \begin{equation}
        \DI(F') = \DI(F) - P(X \in Q(c_G)) / P(X \in \Bcal)
    \end{equation}
    if group $b$ is still disadvantaged in $F'$ (i.e. $\DI(F) \geq P(X \in Q(c_G)) / P(X \in \Bcal)$), and otherwise
    \begin{equation}\label{eq:unaware_LR_di_ineq}
        | \DI(F) - \DI(F') | \leq P(X \in Q(c_G)) / P(X \in \Bcal).
    \end{equation}
\end{proposition}

The focus on one specific unaware model provides a loose but already interesting lower bound on the possible reduction of DI:
As a quantitative example, consider the ACS Income dataset discussed in the following section. In this setting, a logistic regression model assigns a coefficient of $-0.86$ for the protected attribute sex.
We observe that the proportion of women is $P(X \in \Bcal) = 0.48$ and $P(X \in Q(c_G)) = 6.9\%$ is the set of women whose predicted probabilities fall within the range $[0.5, 0.70)$, corresponding to the sigmoid of $0.86$, i.e., $\sigma(0.86) = 0.70$.
If we were to flip decisions in this region, the accuracy would change by at most $0.7$ percentage points either down or up, while the disparate impact would decrease by at least $14.5$ percentage points.\footnote{With a base accuracy of 77,4\% and a DI of 0.052, this translates to a relative accuracy change of 0.009 vs a relative reduction in DI of 0.64---see Figure~\ref{fig:scatter_results}.\label{fn:acs_inc_bound}}

Note that this result only represents one possible model within the $(2 \sigma(-c_{G}) \cdot P(X \in Q(c_G)))$-Rashomon set, and therefore serves as a lower bound on the achievable $DI$ reduction.
It is unlikely to be optimal since it optimises the other coefficients `under the assumption' that $c_G$ will be used; it also only modifies the predictions for women, leaving the men's unchanged.
In summary, we have provided general upper bounds on disparate impact reduction within some $\eps$-Rashomon set that are achievable for unconstrained model classes and we have a na\"ive lower bound for logistic regression.
In the next section, we empirically investigate how unaware and aware models compare on real datasets.






 

    


\section{Empirical findings}
\label{s:empirical}

After the theoretical results above, we now empirically analyse the effect of Fairness through Unawareness (FtU) on Disparate Impact (DI) and Accuracy.
We consider three datasets, with different model classes and protected attributes.
All code, also for Section~\ref{ss:aware_LR} above, is available at \url{https://github.com/ben-hoeltgen/ftu-mm/}.

\subsection{Setup}
\label{ss:empirical_setup}

\textbf{1. Models:} We analyse the behaviour of two model classes, logistic regression (LR) and
gradient-boosted decision trees (GBM), using the popular \texttt{scikit-learn} package (more details in Appendix~\ref{app:models}).

\noindent \textbf{2. Data:} We perform experiments on three datasets: 
The ACS Income and ACS Employment datasets from the \texttt{folktables} package \cite{ding2021} from the US Census and the active labour market policy (ALMP) evaluation dataset from Switzerland \cite{lechner2020}.
Table \ref{tab:datasets} in Appendix~\ref{app:datasets} contains a summary of the data. 
In ACS Income and ACS Employment, the target variables are whether a person's income is above 50k and whether they are employed, respectively.
We consider two binary protected attributes: sex (male/female) and race (white/black). 
Note that the US Census asks for sex rather than gender, which is reflected in the derived dataset.
Race is self-indicated in the census, and we restrict to the white/black categories.\footnote{We want to emphasise that race categories are an over-simplification of the complex notion of racialization. Their use should always be contextualised and avoided where possible \cite{doh2025}.}
In the ALMP dataset, our target variable is whether a person will be employed for at least 6 months within the next two years.
We consider two binary protected attributes: gender (male/female) and citizenship (Swiss/foreign). 
In line with previous literature \cite{zezulka2024}, we exclude caseworker information and restrict data to German-speaking cantons.

\subsection{Results}

\begin{figure}[h]
    \vspace{-10pt}
    \centering
    \includegraphics[width=\linewidth]{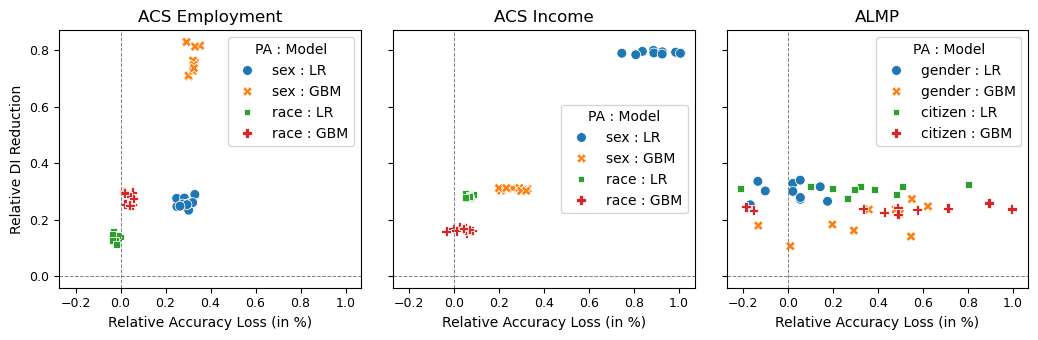}
    \caption{Relative reduction in Accuracy vs in Disparate Impact (DI) across three datasets, with two model classes (LR: logistic regression and GBM: gradient boosted trees) and two protected attributes (PAs) each. Each dot corresponds to the difference in Accuracy and DI between the aware and the corresponding unaware model trained on the same random train/test split. Note how the unaware model achieves a reduction in accuracy which is always below 1\% and sometimes it even increases, while yielding a reduction in DI that is much more substantial (between $15-80\%$).}
    \label{fig:scatter_results}
\end{figure}

The main results, shown in Figure~\ref{fig:scatter_results}, indicate that Accuracy remains similar between aware and unaware models (\emph{i.e.}, models that exclude the protected attribute), while DI is often significantly reduced in the unaware models, yielding fairer models with similar levels of accuracy.\footnote{This result is consistent with previous analysis on the German unemployment data, where it was observed that ``in- or excluding protected attributes has little effect on overall performance'' \cite{kern2024}. However, this observation is not further discussed there.}
The results on the ALMP dataset show general multiplicity, whereas the results on the ACS datasets exhibit less multiplicity. This is not surprising, given that the number of data points is much larger in the ACS datasets than in the ALMP dataset (Table~\ref{tab:datasets}).
The fact that the observed results hold on such large datasets suggests that the effect does not simply disappear with more data.

\begin{figure}[h]
    \centering
    \includegraphics[width=\linewidth]{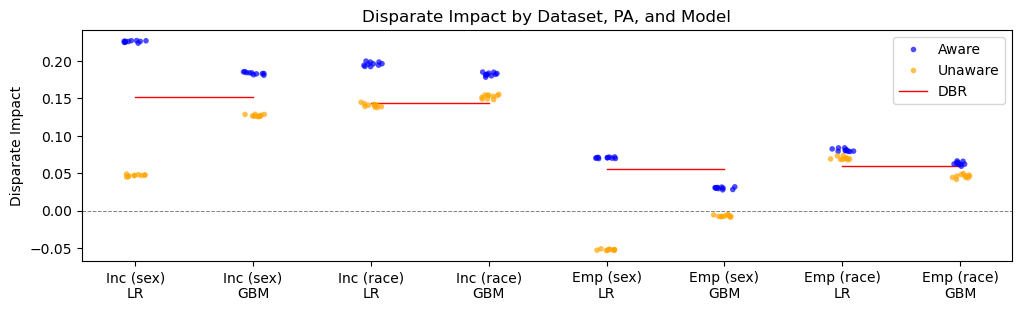}
    \caption{Disparate impact for the aware and unaware models in different settings. The red lines indicate the difference in base rates in the data. Note how the unaware models achieve lower disparate impact than the aware models.}
    \label{fig:acs_dbr}
\end{figure}

When considering race as the protected attribute in the ACS datasets, the reduction in disparate impact is generally smaller than with sex, but reaches around 20\% across all settings at virtually no loss in accuracy.
For settings where sex is the protected attribute, there is an interesting difference in the results on the ACS Income vs the ACS Employment datasets. On ACS Income, the largest reduction in DI ($\approx 80\%$) is obtained with the unaware LR model (consistent with, but slightly exceeding, the lower bound of Section~\ref{ss:aware_LR}, see footnote~\ref{fn:acs_inc_bound}). In contrast, on ACS Employment, the largest reduction of DI (also $\approx 80\%$) corresponds to the unaware GBM. 
This result can be explained by two effects that are observable in Figure~\ref{fig:acs_dbr}.
First, aware LR can lead to particularly high disparate impact, as discussed in Section~\ref{ss:aware_LR}. 
This effect is particularly pronounced when sex is the protected attribute. For instance, in the case of the ACS Income dataset, the LR model leads to a disparate impact that far exceeds the base rate difference in the data (indicated by the red line). 
Second, in the ACS Employment dataset, the LR model with sex as the protected attribute actually results in higher positive classification rates for women than for men, suggesting that women may have stronger predictive features for employment than men. As a result,  the change in disparity between men and women is more noticeable with LR than with GBM (see Figure~\ref{fig:acs_dbr}), even though the absolute DI values change less (see Figure~\ref{fig:scatter_results}).
This difference between ACS Employment and ACS Income can be attributed to the lower base rate differences in the former (see Table~\ref{tab:datasets}).

\section{Discussion: Real-world use case}
\label{s:ams}

We have shown that including protected attributes like sex or gender in models can lead to substantially less equitable classifications, often without meaningful gains in accuracy---and sometimes without improving accuracy at all. 
However, what ultimately matters is not just how models classify individuals, but the real-world impact that these systems have when used to guide decisions. In other words, what matters are the \textit{policies} that these models inform \cite{kuppler2022}.
As others have pointed out, \emph{``[i]t is policies and their effects that are just or unjust; `fair' predictors can both support unjust policies and undermine just policy"} \cite[p. 1984]{zezulka2024}.
Machine learning models are increasingly used in resource allocation, where limited goods or opportunities---like counselling or training programs---are distributed based on risk predictions, often through threshold-based policies.\footnote{While our analyses have focused on the 50\% threshold, we show below that the underlying insights apply more broadly.} 
An important question is whether predictions are even helpful in such contexts \cite{shirali2024, perdomo2023, perdomo2024}.
Furthermore, there is the concern that predicting risk irrespective of treatment may not suffice since this risk does not necessarily correlate with treatment effectiveness \cite{zezulka2024, sharma2025}.
With these caveats, we now turn to examine the impact of Fairness through Unawareness in a real-world application that has attracted considerable attention in Europe in recent years: the statistical profiling of job seekers, 
which is increasingly used in labour market programs across the world to support decisions on how to allocate resources such as job training opportunities \cite{desiere2019}.
These systems aim to predict an individual's risk of long-term unemployment such that costly services can be targeted more efficiently, ideally to those for whom the intervention will have the greatest positive impact. 
Different models use different predictors. For example, gender has been explicitly included in profiling models used in Australia \cite{lipp2005} and Austria \cite{holl2018}. 

A recent example is the Austrian AMS algorithm, which uses an undisclosed stratification procedure \cite{gamper2020} that is approximated by a published logistic regression model \cite{holl2018}, and includes gender as a feature.
The AMS algorithm has sparked significant legal and political controversy, leading to multiple revisions and delays in its deployment. As a result, it has not yet been implemented in practice. Critics argue that the model creates a false impression of objectivity and has been specifically accused of reinforcing gender discrimination \cite{allhutter2020}.
The Austrian system stands out by targeting individuals of medium risk: the model categorizes job seekers into three groups---low (A), medium (B) and high (C) risk of long-term unemployment---but only those in the medium-risk group (B) are offered access to the costly job training programs.
There are various issues with this approach, most notably its reliance on observational data to make causal claims 
\cite{zezulka2024, sharma2025}, such as the assumption that individuals at medium risk benefit the most from the intervention. 
In line with the broader theme of this paper, we highlight another common but flawed assumption: that using more attributes necessarily leads to more objective (and hence fairer) decisions.
One major criticism of the AMS algorithm has been its discriminatory impact, especially with regard to gender. The model assigned a negative coefficient of $-0.14$ to women and applied an additional penalty for care obligations, a variable that is only present for women. As a result, 5\% of women were placed in the high-risk group C---deemed ineligible for job training---when compared to only 3\% of men \cite{holl2018}.
Defenders of the system argued that its predictions simply reflect actual labour market conditions, stating it follows \emph{``the real chances on the labour market as far as possible''} (as cited in \citet{allhutter2020}).


\begin{table}[h]
    \centering
    \begin{tabular}{llcccc}
        \toprule
        \textbf{Model} & \textbf{Mode} & \textbf{Male} & \textbf{Female} & $\mathbf{\Delta} \downarrow$ & \textbf{AUROC} $\uparrow$ \\
        \midrule
        \multirow{2}{*}{\textbf{LR}} & Aware & $1.6\% \pm 0.1\%$ & $2.2\% \pm 0.2\%$ & $0.7\% \pm 0.3\%$ & \textbf{0.680} $\pm 0.002$ \\
        & Unaware & $1.7\% \pm 0.1\%$ & $2.0\% \pm 0.2\%$ & \textbf{0.2\%} $\pm 0.2\%$ & \textbf{0.680} $\pm 0.002$ \\
        \midrule
        \multirow{2}{*}{\textbf{GBM}} & Aware & $3.9\% \pm 0.3\%$ & $8.3\% \pm 0.4\%$ & $4.4\% \pm 0.4\%$ & \textbf{0.700} $\pm 0.003$ \\
        & Unaware & $4.5\% \pm 0.1\%$ & $6.1\% \pm 0.2\%$ & \textbf{1.6}\% $\pm 0.2\%$ & $0.697 \pm 0.003$ \\
        \bottomrule
    \end{tabular}
    \caption{Group C sizes for both the aware and unaware LR and GBM models on the original ALMP dataset (means and STDs over 10 random splits). The $\Delta$ column reflects the difference between men and women. Note how the unaware models yield more equitable results with negligible loss in accuracy. Best results indicated in bold font.}
    \label{tab:almp}
\end{table}

However, as shown in previous sections, there is no direct or reliable link between observed label rates (or supposed `real chances') and the disparities in outcomes produced by model-based decisions. 
In the following analysis, we reconstruct the AMS setting as faithfully as possible (given the lack of access to the original data) and explore the effects of excluding gender on both accuracy and disparate impact.
Lacking the original data, we base our analysis on the active labour market policy (ALMP) dataset from Switzerland \cite{lechner2020}, focusing on the German-speaking cantons. Although the features in this dataset differ somewhat from those used in the AMS model, we aim to approximate the setting, not replicate it exactly. 
We adopt the same prediction target used by the AMS algorithm to determine assignment to group C, namely, whether an individual is employed for at least 6 out of the next 24 months.\footnote{A second model with a slightly different target is used to determine assignment to group A.}
We train both LR and GBM models and present the results in Table~\ref{tab:almp}.
In our analysis, we focus on two main outcomes: (1) the per-gender proportion of individuals assigned to group C (those considered too high-risk to receive job training), and the difference between genders; and (2) the AUROC of the models, given that the decision criterion is the 25\% threshold.
As reflected in Table~\ref{tab:almp}, while omitting gender has little impact on AUROC, it leads to a more balanced assignment between genders, \emph{i.e.}, the proportion of individuals deemed too risky is more equitable in the unaware model. 
Hence, even if there is no straightforward connection between AUROC and efficiency, these results suggest that Fairness through Unawareness can lead to more equitable but equally efficient policies in practice.


\section{Conclusion}
\label{s:conclusion}

In this paper, we have shown that unaware models, which omit a protected attribute, can significantly reduce disparate impact across groups defined by that attribute while maintaining accuracy. Our theoretically and empirical analysis suggests that this effect is common, at least on tabular data.
Our findings urge caution when using logistic regression, as it can produce unintended disparities.
Note that our analyses do not suggest that LR or protected attributes should never be used (although there can be legal constraints), nor that FtU is sufficient for ensuring fair algorithmic decision-making. 
Indeed, we did not compare FtU to fairness interventions, such as threshold adjustments, precisely because our point is less about achieving SOTA results than about demonstrating the importance of more basic modelling choices (even though such a comparison may be of interest for the intervention perspective).
In this sense, our results underscore the need for critical evaluation of algorithmic choices as part of the broader deployment pipeline of machine learning models \cite{black2024}.
More generally, our results challenge the notion of a strict fairness-accuracy trade-off, which has been theoretically suggested based on true distributions and Bayes-optimal predictors \cite{corbett2017,menon2018}.
We argue that the relevance of such theoretical claims to real-world applications should be scrutinised, given the potential dangers of uncritically applying idealised assumptions to real-world problems with consequential impact on people's lives \cite{holtgen2024position}.



\section*{Ethical considerations}

A potential adverse effect of this work is that it could induce model designers to exclude protected attributes by default, even in cases where the benefits may outweigh the harms, as suggested to be the case for some health settings \cite{corbett2017}. However, our work does not warrant such a conclusion and we generally suggest investigating both unaware and aware models where possible. Another unwarranted inference would be to claim that unawareness is enough for ensuring fairness or equality –- this conclusion is also not supported by our work; we merely show that it can be a useful means toward this end.


\begin{acks}
BH is supported by the German Federal Ministry of Education and Research (BMBF): Tübingen AI Center, FKZ: 01IS18039A. BH also acknowledges support by the International Max Planck Research School for Intelligent Systems (IMPRS-IS) and travel support from ELIAS (GA no 101120237).

NO is supported by a nominal grant received at the ELLIS Unit Alicante Foundation from the Regional Government of Valencia in Spain (Convenio Singular signed with Generalitat Valenciana, Conselleria de Innovaci\'on, Industria, Comercio y Turismo, Direcci\'on General de Innovaci\'on) and by EU - HE ELIAS – Grant Agreement 101120237. Views and opinions expressed are however those of the author(s) only and do not necessarily reflect
those of the European Union or the European Health and Digital Executive Agency (HaDEA).  

We would also like to thank Adrián Arnaiz-Rodríguez for insightful discussions throughout the project and Sebastian Zezulka as well as the anonymous reviewers for helpful feedback on the manuscript.
\end{acks}

\bibliographystyle{ACM-Reference-Format}
\bibliography{library}


\begin{thebibliography}{46}


\ifx \showCODEN    \undefined \def \showCODEN     #1{\unskip}     \fi
\ifx \showISBNx    \undefined \def \showISBNx     #1{\unskip}     \fi
\ifx \showISBNxiii \undefined \def \showISBNxiii  #1{\unskip}     \fi
\ifx \showISSN     \undefined \def \showISSN      #1{\unskip}     \fi
\ifx \showLCCN     \undefined \def \showLCCN      #1{\unskip}     \fi
\ifx \shownote     \undefined \def \shownote      #1{#1}          \fi
\ifx \showarticletitle \undefined \def \showarticletitle #1{#1}   \fi
\ifx \showURL      \undefined \def \showURL       {\relax}        \fi
\providecommand\bibfield[2]{#2}
\providecommand\bibinfo[2]{#2}
\providecommand\natexlab[1]{#1}
\providecommand\showeprint[2][]{arXiv:#2}

\bibitem[Allhutter et~al\mbox{.}(2020)]%
        {allhutter2020}
\bibfield{author}{\bibinfo{person}{Doris Allhutter}, \bibinfo{person}{Florian Cech}, \bibinfo{person}{Fabian Fischer}, \bibinfo{person}{Gabriel Grill}, {and} \bibinfo{person}{Astrid Mager}.} \bibinfo{year}{2020}\natexlab{}.
\newblock \showarticletitle{Algorithmic profiling of job seekers in {Austria}: how austerity politics are made effective}.
\newblock \bibinfo{journal}{\emph{Frontiers in Big Data}} (\bibinfo{year}{2020}), \bibinfo{pages}{5}.
\newblock


\bibitem[Bach et~al\mbox{.}(2023)]%
        {bach2023}
\bibfield{author}{\bibinfo{person}{Ruben~L Bach}, \bibinfo{person}{Christoph Kern}, \bibinfo{person}{Hannah Mautner}, {and} \bibinfo{person}{Frauke Kreuter}.} \bibinfo{year}{2023}\natexlab{}.
\newblock \showarticletitle{The impact of modeling decisions in statistical profiling}.
\newblock \bibinfo{journal}{\emph{Data \& Policy}}  \bibinfo{volume}{5} (\bibinfo{year}{2023}), \bibinfo{pages}{e32}.
\newblock


\bibitem[Barocas et~al\mbox{.}(2023)]%
        {fairmlbook}
\bibfield{author}{\bibinfo{person}{Solon Barocas}, \bibinfo{person}{Moritz Hardt}, {and} \bibinfo{person}{Arvind Narayanan}.} \bibinfo{year}{2023}\natexlab{}.
\newblock \bibinfo{booktitle}{\emph{Fairness and Machine Learning: Limitations and Opportunities}}.
\newblock \bibinfo{publisher}{MIT Press}.
\newblock


\bibitem[Barocas and Selbst(2016)]%
        {barocas2016}
\bibfield{author}{\bibinfo{person}{Solon Barocas} {and} \bibinfo{person}{Andrew~D Selbst}.} \bibinfo{year}{2016}\natexlab{}.
\newblock \showarticletitle{Big data's disparate impact}.
\newblock \bibinfo{journal}{\emph{California Law Review}}  \bibinfo{volume}{104} (\bibinfo{year}{2016}), \bibinfo{pages}{671}.
\newblock


\bibitem[Black et~al\mbox{.}(2024a)]%
        {black2024}
\bibfield{author}{\bibinfo{person}{Emily Black}, \bibinfo{person}{John~Logan Koepke}, \bibinfo{person}{Pauline Kim}, \bibinfo{person}{Solon Barocas}, {and} \bibinfo{person}{Mingwei Hsu}.} \bibinfo{year}{2024}\natexlab{a}.
\newblock \showarticletitle{Less discriminatory algorithms}.
\newblock \bibinfo{journal}{\emph{Georgetown Law Journal}} \bibinfo{volume}{113}, \bibinfo{number}{1} (\bibinfo{year}{2024}).
\newblock


\bibitem[Black et~al\mbox{.}(2024b)]%
        {black2024short}
\bibfield{author}{\bibinfo{person}{Emily Black}, \bibinfo{person}{Logan Koepke}, \bibinfo{person}{Pauline Kim}, \bibinfo{person}{Solon Barocas}, {and} \bibinfo{person}{Mingwei Hsu}.} \bibinfo{year}{2024}\natexlab{b}.
\newblock \showarticletitle{The legal duty to search for less discriminatory algorithms}.
\newblock \bibinfo{journal}{\emph{arXiv preprint arXiv:2406.06817}} (\bibinfo{year}{2024}).
\newblock


\bibitem[Black et~al\mbox{.}(2022)]%
        {black2022}
\bibfield{author}{\bibinfo{person}{Emily Black}, \bibinfo{person}{Manish Raghavan}, {and} \bibinfo{person}{Solon Barocas}.} \bibinfo{year}{2022}\natexlab{}.
\newblock \showarticletitle{Model multiplicity: Opportunities, concerns, and solutions}. In \bibinfo{booktitle}{\emph{Proceedings of the 2022 ACM Conference on Fairness, Accountability, and Transparency}}. \bibinfo{pages}{850--863}.
\newblock


\bibitem[Boyd and Crawford(2012)]%
        {boyd2012}
\bibfield{author}{\bibinfo{person}{Danah Boyd} {and} \bibinfo{person}{Kate Crawford}.} \bibinfo{year}{2012}\natexlab{}.
\newblock \showarticletitle{Critical questions for big data: Provocations for a cultural, technological, and scholarly phenomenon}.
\newblock \bibinfo{journal}{\emph{Information, Communication \& Society}} \bibinfo{volume}{15}, \bibinfo{number}{5} (\bibinfo{year}{2012}), \bibinfo{pages}{662--679}.
\newblock


\bibitem[Breiman(2001)]%
        {breiman2001}
\bibfield{author}{\bibinfo{person}{Leo Breiman}.} \bibinfo{year}{2001}\natexlab{}.
\newblock \showarticletitle{Statistical modeling: The two cultures (with comments and a rejoinder by the author)}.
\newblock \bibinfo{journal}{\emph{Statistical science}} \bibinfo{volume}{16}, \bibinfo{number}{3} (\bibinfo{year}{2001}), \bibinfo{pages}{199--231}.
\newblock


\bibitem[Calders and Verwer(2010)]%
        {calders2010}
\bibfield{author}{\bibinfo{person}{Toon Calders} {and} \bibinfo{person}{Sicco Verwer}.} \bibinfo{year}{2010}\natexlab{}.
\newblock \showarticletitle{Three naive bayes approaches for discrimination-free classification}.
\newblock \bibinfo{journal}{\emph{Data mining and knowledge discovery}}  \bibinfo{volume}{21} (\bibinfo{year}{2010}), \bibinfo{pages}{277--292}.
\newblock


\bibitem[Chouldechova(2017)]%
        {chouldechova2017}
\bibfield{author}{\bibinfo{person}{Alexandra Chouldechova}.} \bibinfo{year}{2017}\natexlab{}.
\newblock \showarticletitle{Fair prediction with disparate impact: A study of bias in recidivism prediction instruments}.
\newblock \bibinfo{journal}{\emph{Big data}} \bibinfo{volume}{5}, \bibinfo{number}{2} (\bibinfo{year}{2017}), \bibinfo{pages}{153--163}.
\newblock


\bibitem[Coots et~al\mbox{.}(2023)]%
        {coots2023}
\bibfield{author}{\bibinfo{person}{Madison Coots}, \bibinfo{person}{Soroush Saghafian}, \bibinfo{person}{David Kent}, {and} \bibinfo{person}{Sharad Goel}.} \bibinfo{year}{2023}\natexlab{}.
\newblock \showarticletitle{Reevaluating the role of race and ethnicity in diabetes screening}.
\newblock \bibinfo{journal}{\emph{arXiv preprint arXiv:2306.10220}} (\bibinfo{year}{2023}).
\newblock


\bibitem[Corbett-Davies et~al\mbox{.}(2023)]%
        {corbett2023}
\bibfield{author}{\bibinfo{person}{Sam Corbett-Davies}, \bibinfo{person}{Johann~D Gaebler}, \bibinfo{person}{Hamed Nilforoshan}, \bibinfo{person}{Ravi Shroff}, {and} \bibinfo{person}{Sharad Goel}.} \bibinfo{year}{2023}\natexlab{}.
\newblock \showarticletitle{The measure and mismeasure of fairness}.
\newblock \bibinfo{journal}{\emph{The Journal of Machine Learning Research}} \bibinfo{volume}{24}, \bibinfo{number}{1} (\bibinfo{year}{2023}), \bibinfo{pages}{14730--14846}.
\newblock


\bibitem[Corbett-Davies et~al\mbox{.}(2017)]%
        {corbett2017}
\bibfield{author}{\bibinfo{person}{Sam Corbett-Davies}, \bibinfo{person}{Emma Pierson}, \bibinfo{person}{Avi Feller}, \bibinfo{person}{Sharad Goel}, {and} \bibinfo{person}{Aziz Huq}.} \bibinfo{year}{2017}\natexlab{}.
\newblock \showarticletitle{Algorithmic decision making and the cost of fairness}. In \bibinfo{booktitle}{\emph{Proceedings of the 23rd acm sigkdd international conference on knowledge discovery and data mining}}. \bibinfo{pages}{797--806}.
\newblock


\bibitem[Coston et~al\mbox{.}(2021)]%
        {coston2021}
\bibfield{author}{\bibinfo{person}{Amanda Coston}, \bibinfo{person}{Ashesh Rambachan}, {and} \bibinfo{person}{Alexandra Chouldechova}.} \bibinfo{year}{2021}\natexlab{}.
\newblock \showarticletitle{Characterizing fairness over the set of good models under selective labels}. In \bibinfo{booktitle}{\emph{International Conference on Machine Learning}}. PMLR, \bibinfo{pages}{2144--2155}.
\newblock


\bibitem[Cruz and Hardt(2024)]%
        {cruz2024}
\bibfield{author}{\bibinfo{person}{Andr{\'e} Cruz} {and} \bibinfo{person}{Moritz Hardt}.} \bibinfo{year}{2024}\natexlab{}.
\newblock \showarticletitle{Unprocessing Seven Years of Algorithmic Fairness}. In \bibinfo{booktitle}{\emph{The Twelfth International Conference on Learning Representations}}.
\newblock


\bibitem[Desiere et~al\mbox{.}(2019)]%
        {desiere2019}
\bibfield{author}{\bibinfo{person}{Sam Desiere}, \bibinfo{person}{Kristine Langenbucher}, {and} \bibinfo{person}{Ludo Struyven}.} \bibinfo{year}{2019}\natexlab{}.
\newblock \showarticletitle{Statistical profiling in public employment services. An international comparison}.
\newblock \bibinfo{journal}{\emph{OECD Social, Employment and Migration Working Papers}} \bibinfo{number}{224} (\bibinfo{year}{2019}).
\newblock


\bibitem[Ding et~al\mbox{.}(2021)]%
        {ding2021}
\bibfield{author}{\bibinfo{person}{Frances Ding}, \bibinfo{person}{Moritz Hardt}, \bibinfo{person}{John Miller}, {and} \bibinfo{person}{Ludwig Schmidt}.} \bibinfo{year}{2021}\natexlab{}.
\newblock \showarticletitle{Retiring adult: New datasets for fair machine learning}.
\newblock \bibinfo{journal}{\emph{Advances in neural information processing systems}}  \bibinfo{volume}{34} (\bibinfo{year}{2021}), \bibinfo{pages}{6478--6490}.
\newblock


\bibitem[Doh et~al\mbox{.}(2025)]%
        {doh2025}
\bibfield{author}{\bibinfo{person}{Miriam Doh}, \bibinfo{person}{Benedikt H{\"o}ltgen}, \bibinfo{person}{Piera Riccio}, {and} \bibinfo{person}{Nuria~M Oliver}.} \bibinfo{year}{2025}\natexlab{}.
\newblock \showarticletitle{Position: The categorization of race in ml is a flawed premise}. In \bibinfo{booktitle}{\emph{Forty-second International Conference on Machine Learning}}.
\newblock


\bibitem[Fazelpour and Lipton(2020)]%
        {fazelpour2020}
\bibfield{author}{\bibinfo{person}{Sina Fazelpour} {and} \bibinfo{person}{Zachary~C Lipton}.} \bibinfo{year}{2020}\natexlab{}.
\newblock \showarticletitle{Algorithmic fairness from a non-ideal perspective}. In \bibinfo{booktitle}{\emph{Proceedings of the AAAI/ACM Conference on AI, Ethics, and Society}}. \bibinfo{pages}{57--63}.
\newblock


\bibitem[Feldman et~al\mbox{.}(2015)]%
        {feldman2015}
\bibfield{author}{\bibinfo{person}{Michael Feldman}, \bibinfo{person}{Sorelle~A Friedler}, \bibinfo{person}{John Moeller}, \bibinfo{person}{Carlos Scheidegger}, {and} \bibinfo{person}{Suresh Venkatasubramanian}.} \bibinfo{year}{2015}\natexlab{}.
\newblock \showarticletitle{Certifying and removing disparate impact}. In \bibinfo{booktitle}{\emph{proceedings of the 21th ACM SIGKDD international conference on knowledge discovery and data mining}}. \bibinfo{pages}{259--268}.
\newblock


\bibitem[Gamper et~al\mbox{.}(2020)]%
        {gamper2020}
\bibfield{author}{\bibinfo{person}{Jutta Gamper}, \bibinfo{person}{G{\"u}nter Kernbei{\ss}}, {and} \bibinfo{person}{Michael Wagner-Pinter}.} \bibinfo{year}{2020}\natexlab{}.
\newblock \bibinfo{title}{Das Assistenzsystem AMAS. Zweck, Grundlagen, Anwendung}.
\newblock


\bibitem[Gillis et~al\mbox{.}(2024)]%
        {gillis2024}
\bibfield{author}{\bibinfo{person}{Talia~B Gillis}, \bibinfo{person}{Vitaly Meursault}, {and} \bibinfo{person}{Berk Ustun}.} \bibinfo{year}{2024}\natexlab{}.
\newblock \showarticletitle{Operationalizing the Search for Less Discriminatory Alternatives in Fair Lending}. In \bibinfo{booktitle}{\emph{The 2024 ACM Conference on Fairness, Accountability, and Transparency}}. \bibinfo{pages}{377--387}.
\newblock


\bibitem[Hardt et~al\mbox{.}(2016)]%
        {hardt2016}
\bibfield{author}{\bibinfo{person}{Moritz Hardt}, \bibinfo{person}{Eric Price}, {and} \bibinfo{person}{Nati Srebro}.} \bibinfo{year}{2016}\natexlab{}.
\newblock \showarticletitle{Equality of opportunity in supervised learning}.
\newblock \bibinfo{journal}{\emph{Advances in neural information processing systems}}  \bibinfo{volume}{29} (\bibinfo{year}{2016}).
\newblock


\bibitem[Holl et~al\mbox{.}(2018)]%
        {holl2018}
\bibfield{author}{\bibinfo{person}{J{\"u}rgen Holl}, \bibinfo{person}{G{\"u}nter Kernbei{\ss}}, {and} \bibinfo{person}{Michael Wagner-Pinter}.} \bibinfo{year}{2018}\natexlab{}.
\newblock \showarticletitle{{Das AMS-Arbeitsmarktchancen-Modell}}.
\newblock \bibinfo{journal}{\emph{Arbeitsmarktservice {\"O}sterreich, Wien}} (\bibinfo{year}{2018}).
\newblock


\bibitem[H{\"o}ltgen and Williamson(2025)]%
        {holtgen2024position}
\bibfield{author}{\bibinfo{person}{Benedikt H{\"o}ltgen} {and} \bibinfo{person}{Robert~C Williamson}.} \bibinfo{year}{2025}\natexlab{}.
\newblock \showarticletitle{We should avoid the assumption of data-generating probability distributions in social settings}.
\newblock \bibinfo{journal}{\emph{arXiv preprint arXiv:2407.17395}} (\bibinfo{year}{2025}).
\newblock


\bibitem[Kern et~al\mbox{.}(2024)]%
        {kern2024}
\bibfield{author}{\bibinfo{person}{Christoph Kern}, \bibinfo{person}{Ruben Bach}, \bibinfo{person}{Hannah Mautner}, {and} \bibinfo{person}{Frauke Kreuter}.} \bibinfo{year}{2024}\natexlab{}.
\newblock \showarticletitle{When small decisions have big impact: fairness implications of algorithmic profiling schemes}.
\newblock \bibinfo{journal}{\emph{ACM Journal on Responsible Computing}} \bibinfo{volume}{1}, \bibinfo{number}{4} (\bibinfo{year}{2024}), \bibinfo{pages}{1--30}.
\newblock


\bibitem[Kuppler et~al\mbox{.}(2022)]%
        {kuppler2022}
\bibfield{author}{\bibinfo{person}{Matthias Kuppler}, \bibinfo{person}{Christoph Kern}, \bibinfo{person}{Ruben~L Bach}, {and} \bibinfo{person}{Frauke Kreuter}.} \bibinfo{year}{2022}\natexlab{}.
\newblock \showarticletitle{From fair predictions to just decisions? Conceptualizing algorithmic fairness and distributive justice in the context of data-driven decision-making}.
\newblock \bibinfo{journal}{\emph{Frontiers in sociology}}  \bibinfo{volume}{7} (\bibinfo{year}{2022}), \bibinfo{pages}{883999}.
\newblock


\bibitem[Lechner et~al\mbox{.}(2020)]%
        {lechner2020}
\bibfield{author}{\bibinfo{person}{Michael Lechner}, \bibinfo{person}{Michael Knaus}, \bibinfo{person}{Martin Huber}, \bibinfo{person}{Markus Frölich}, \bibinfo{person}{Stefanie Behncke}, \bibinfo{person}{Giovanni Mellace}, {and} \bibinfo{person}{Anthony Strittmatter}.} \bibinfo{year}{2020}\natexlab{}.
\newblock \showarticletitle{Swiss Active Labor Market Policy Evaluation [Dataset]}.
\newblock \bibinfo{journal}{\emph{Distributed by FORS}} (\bibinfo{year}{2020}).
\newblock
\href{https://doi.org/10.23662/FORS-DS-1203-1}{doi:\nolinkurl{10.23662/FORS-DS-1203-1}}


\bibitem[Lipp(2005)]%
        {lipp2005}
\bibfield{author}{\bibinfo{person}{Robert Lipp}.} \bibinfo{year}{2005}\natexlab{}.
\newblock \showarticletitle{Job Seeker Profiling}.
\newblock \bibinfo{journal}{\emph{The Australian Experience}} (\bibinfo{year}{2005}).
\newblock


\bibitem[Marx et~al\mbox{.}(2020)]%
        {marx2020}
\bibfield{author}{\bibinfo{person}{Charles Marx}, \bibinfo{person}{Flavio Calmon}, {and} \bibinfo{person}{Berk Ustun}.} \bibinfo{year}{2020}\natexlab{}.
\newblock \showarticletitle{Predictive multiplicity in classification}. In \bibinfo{booktitle}{\emph{International Conference on Machine Learning}}. PMLR, \bibinfo{pages}{6765--6774}.
\newblock


\bibitem[Menon and Williamson(2018)]%
        {menon2018}
\bibfield{author}{\bibinfo{person}{Aditya~Krishna Menon} {and} \bibinfo{person}{Robert~C Williamson}.} \bibinfo{year}{2018}\natexlab{}.
\newblock \showarticletitle{The cost of fairness in binary classification}. In \bibinfo{booktitle}{\emph{Conference on Fairness, accountability and transparency}}. PMLR, \bibinfo{pages}{107--118}.
\newblock


\bibitem[Moss(2022)]%
        {moss2022}
\bibfield{author}{\bibinfo{person}{Emanuel Moss}.} \bibinfo{year}{2022}\natexlab{}.
\newblock \showarticletitle{The objective function: Science and society in the age of machine intelligence}.
\newblock \bibinfo{journal}{\emph{arXiv:2209.10418}} (\bibinfo{year}{2022}).
\newblock


\bibitem[Pedreshi et~al\mbox{.}(2008)]%
        {pedreshi2008}
\bibfield{author}{\bibinfo{person}{Dino Pedreshi}, \bibinfo{person}{Salvatore Ruggieri}, {and} \bibinfo{person}{Franco Turini}.} \bibinfo{year}{2008}\natexlab{}.
\newblock \showarticletitle{Discrimination-aware data mining}. In \bibinfo{booktitle}{\emph{Proceedings of the 14th ACM SIGKDD International Conference on Knowledge Discovery and Data Mining}}. \bibinfo{pages}{560--568}.
\newblock


\bibitem[Perdomo(2024)]%
        {perdomo2024}
\bibfield{author}{\bibinfo{person}{Juan~Carlos Perdomo}.} \bibinfo{year}{2024}\natexlab{}.
\newblock \showarticletitle{The relative value of prediction in algorithmic decision making}. In \bibinfo{booktitle}{\emph{Proceedings of the 41st International Conference on Machine Learning}}. \bibinfo{pages}{40439--40460}.
\newblock


\bibitem[Perdomo et~al\mbox{.}(2023)]%
        {perdomo2023}
\bibfield{author}{\bibinfo{person}{Juan~C Perdomo}, \bibinfo{person}{Tolani Britton}, \bibinfo{person}{Moritz Hardt}, {and} \bibinfo{person}{Rediet Abebe}.} \bibinfo{year}{2023}\natexlab{}.
\newblock \showarticletitle{Difficult Lessons on Social Prediction from {Wisconsin} Public Schools}.
\newblock \bibinfo{journal}{\emph{arXiv:2304.06205}} (\bibinfo{year}{2023}).
\newblock


\bibitem[Porter(1995)]%
        {porter1995}
\bibfield{author}{\bibinfo{person}{Theodore~M Porter}.} \bibinfo{year}{1995}\natexlab{}.
\newblock \bibinfo{booktitle}{\emph{Trust in numbers}}.
\newblock \bibinfo{publisher}{Princeton University Press}.
\newblock


\bibitem[Raghavan and Kim(2024)]%
        {raghavan2024}
\bibfield{author}{\bibinfo{person}{Manish Raghavan} {and} \bibinfo{person}{Pauline~T Kim}.} \bibinfo{year}{2024}\natexlab{}.
\newblock \showarticletitle{Limitations of the" Four-Fifths Rule" and Statistical Parity Tests for Measuring Fairness}.
\newblock \bibinfo{journal}{\emph{Geo. L. Tech. Rev.}}  \bibinfo{volume}{8} (\bibinfo{year}{2024}), \bibinfo{pages}{93}.
\newblock


\bibitem[Rudin et~al\mbox{.}(2024)]%
        {rudin2024}
\bibfield{author}{\bibinfo{person}{Cynthia Rudin}, \bibinfo{person}{Chudi Zhong}, \bibinfo{person}{Lesia Semenova}, \bibinfo{person}{Margo Seltzer}, \bibinfo{person}{Ronald Parr}, \bibinfo{person}{Jiachang Liu}, \bibinfo{person}{Srikar Katta}, \bibinfo{person}{Jon Donnelly}, \bibinfo{person}{Harry Chen}, {and} \bibinfo{person}{Zachery Boner}.} \bibinfo{year}{2024}\natexlab{}.
\newblock \showarticletitle{Amazing things come from having many good models}.
\newblock \bibinfo{journal}{\emph{arXiv preprint arXiv:2407.04846}} (\bibinfo{year}{2024}).
\newblock


\bibitem[Selbst et~al\mbox{.}(2019)]%
        {selbst2019}
\bibfield{author}{\bibinfo{person}{Andrew~D Selbst}, \bibinfo{person}{Danah Boyd}, \bibinfo{person}{Sorelle~A Friedler}, \bibinfo{person}{Suresh Venkatasubramanian}, {and} \bibinfo{person}{Janet Vertesi}.} \bibinfo{year}{2019}\natexlab{}.
\newblock \showarticletitle{Fairness and abstraction in sociotechnical systems}. In \bibinfo{booktitle}{\emph{Proceedings of the conference on fairness, accountability, and transparency}}. \bibinfo{pages}{59--68}.
\newblock


\bibitem[Semenova et~al\mbox{.}(2023)]%
        {semenova2023}
\bibfield{author}{\bibinfo{person}{Lesia Semenova}, \bibinfo{person}{Harry Chen}, \bibinfo{person}{Ronald Parr}, {and} \bibinfo{person}{Cynthia Rudin}.} \bibinfo{year}{2023}\natexlab{}.
\newblock \showarticletitle{A path to simpler models starts with noise}.
\newblock \bibinfo{journal}{\emph{Advances in Neural Information Processing Systems}}  \bibinfo{volume}{36} (\bibinfo{year}{2023}).
\newblock


\bibitem[Sen(2010)]%
        {sen2010}
\bibfield{author}{\bibinfo{person}{Amartya Sen}.} \bibinfo{year}{2010}\natexlab{}.
\newblock \bibinfo{booktitle}{\emph{The idea of justice}}.
\newblock \bibinfo{publisher}{Penguin books}.
\newblock


\bibitem[Sharma and Wilder(2025)]%
        {sharma2025}
\bibfield{author}{\bibinfo{person}{Vibhhu Sharma} {and} \bibinfo{person}{Bryan Wilder}.} \bibinfo{year}{2025}\natexlab{}.
\newblock \showarticletitle{Comparing Targeting Strategies for Maximizing Social Welfare with Limited Resources}.
\newblock \bibinfo{journal}{\emph{The Thirteenth International Conference on Learning Representations}} (\bibinfo{year}{2025}).
\newblock


\bibitem[Shirali et~al\mbox{.}(2024)]%
        {shirali2024}
\bibfield{author}{\bibinfo{person}{Ali Shirali}, \bibinfo{person}{Rediet Abebe}, {and} \bibinfo{person}{Moritz Hardt}.} \bibinfo{year}{2024}\natexlab{}.
\newblock \showarticletitle{Allocation Requires Prediction Only if Inequality Is Low}. In \bibinfo{booktitle}{\emph{Proceedings of the 41st International Conference on Machine Learning}}. \bibinfo{pages}{45114--45153}.
\newblock


\bibitem[Wang et~al\mbox{.}(2022)]%
        {wang2022}
\bibfield{author}{\bibinfo{person}{Angelina Wang}, \bibinfo{person}{Sayash Kapoor}, \bibinfo{person}{Solon Barocas}, {and} \bibinfo{person}{Arvind Narayanan}.} \bibinfo{year}{2022}\natexlab{}.
\newblock \bibinfo{title}{Against predictive optimization: On the legitimacy of decision-making algorithms that optimize predictive accuracy}.
\newblock
\urldef\tempurl%
\url{https://papers.ssrn.com/abstract=4238015}
\showURL{%
\tempurl}


\bibitem[Zezulka and Genin(2024)]%
        {zezulka2024}
\bibfield{author}{\bibinfo{person}{Sebastian Zezulka} {and} \bibinfo{person}{Konstantin Genin}.} \bibinfo{year}{2024}\natexlab{}.
\newblock \showarticletitle{From the Fair Distribution of Predictions to the Fair Distribution of Social Goods: Evaluating the Impact of Fair Machine Learning on Long-Term Unemployment}. In \bibinfo{booktitle}{\emph{The 2024 ACM Conference on Fairness, Accountability, and Transparency}}. \bibinfo{pages}{1984--2006}.
\newblock


\end{thebibliography}


\appendix


\section{More details on experimental setups}

\subsection{Datasets}
\label{app:datasets}

\begin{table}[ht]
    \centering
    \begin{tabular}{ll}
        \toprule
        \textbf{ACS Employment} & \textbf{ACS Income} \\
        \midrule
        SCHL: Education & SCHL: Education \\
        MAR: Marital status & MAR: Marital status \\
        AGEP: Age & AGEP: Age \\
        RAC1P: Race & RAC1P: Race \\
        RELP: Relationship & RELP: Relationship \\
        SEX: Sex & SEX: Sex \\
        DEYE: Vision difficulty & \\
        ANC: Ancestry & \\
        DREM: Cognitive difficulty & \\
        MIG: Move in last year & \\
        CIT: Citizenship & \\
        DIS: Disability & \\
        DEAR: Hearing difficulty & \\
        MIL: Military service & \\
        NATIVITY: Born in US & \\
        ESP: Employment status of parents & \\
         & OCCP: Occupation \\
         & WKHP: Work hours per week\\
         & COW: Class of worker \\
         & POBP: Place of birth \\
        \bottomrule
    \end{tabular}
    \caption{Features used in ACS Employment and Income.}
    \label{tab:features}
\end{table}

\begin{table}[ht]
    \centering
    \begin{tabular}{lcccccr}
        \toprule
        \textbf{Dataset} & \textbf{\# Features} & \textbf{PA} & \textbf{PA Values} & \textbf{Group Size} & \textbf{DBR} & \textbf{\# Datapoints} \\
        \midrule
        \multirow{2}{*}{\textbf{ACS Income}} & \multirow{2}{*}{9/10} & Sex & male/female & 52.1\% & 0.15 & 1,664,500 \\
         \cmidrule(lr){3-7}
         & & Race & white/black & 89.8\% & 0.14 & 1,445,699 \\
        \midrule
        \multirow{2}{*}{\textbf{ACS Employment}} & \multirow{2}{*}{15/16} & Sex & male/female & 49.0\% & 0.06 & 3,236,107 \\
        \cmidrule(lr){3-7}
         & & Race & white/black & 88.9\% & 0.06 & 2,796,670 \\
        \midrule
        \multirow{2}{*}{\textbf{ALMP}} & \multirow{2}{*}{28/29} & Gender & male/female & 56.2\% & 0.04 & 69,372 \\
        \cmidrule(lr){3-7}
         & & Citizenship & Swiss/foreign & 64.3\% & 0.17 & 69,372 \\
        \bottomrule
    \end{tabular}
    \caption{Datasets summary. The number of features differs by one (protected attribute) between aware and unaware models. The `Group Size' column indicates the proportion of data points belonging to the advantaged group, and the 'DBR' column contains the Disparate Benefit Ratio, which is the ratio of favourable outcomes received by the disadvantaged group compared to the advantaged group, indicating the degree of inequality in model predictions across protected groups. When this ratio is much smaller than 1, it indicates disparate treatment.}
    \label{tab:datasets}
\end{table}

The Swiss ALMP dataset features include education, demographics (gender, marital status, age), nationality, region, three related to employment history, two macroeconomic indicators, eight features on previous job characteristics, and ten features on completed job training programs.
More information can be found on the dataset website \cite{lechner2020}.
The features used in the ACS datasets are listed in Table~\ref{tab:features}.
For more details, we refer to the original publication \cite{ding2021}.

\subsection{Models}
\label{app:models}

We use default settings \texttt{LogisticRegression} in \texttt{scikit-learn} but tune \texttt{GradientBoostingClassifier} for higher accuracy.
The parameters are \texttt{n\_estimators=500}, \texttt{max\_depth=10}, \texttt{max\_leaf\_nodes=50}, \texttt{min\_samples\_leaf=500}, very similar to the optimal parameters on the ACS datasets according to previous work (\cite{cruz2024}, personal communication).
We use the same settings for the ALMP experiments.
The accuracies for the default aware models are reported in Table~\ref{tab:abs_accuracies} for comparability.

\begin{table}[ht]
    \centering
    \begin{tabular}{lccccc}
        \toprule
        \textbf{Model} & \textbf{ACS Inc (sex)} & \textbf{ACS Inc (race)} & \textbf{ACS Empl (sex)} & \textbf{ACS Empl (race)} & \textbf{ALMP} \\
        \midrule
        \textbf{LR}  & $77.4\% \pm 0.1\%$ & $77.1\% \pm 0.0\%$ & $77.9\% \pm 0.0\%$ & $78.1\% \pm 0.1\%$ & $65.0\% \pm 0.4\%$ \\
        \textbf{GBM} & $82.1\% \pm 0.1\%$ & $81.8\% \pm 0.1\%$ & $83.2\% \pm 0.1\%$ & $83.2\% \pm 0.0\%$ & $66.2\% \pm 0.3\%$ \\
        \bottomrule
    \end{tabular}
    \caption{Accuracies per setting for the aware model with standard deviations. The accuracies of aware models on ACS datasets differ between the sex and the race setting since we only use a subset of the data in the latter, restricting to `white' and `black'.}
    \label{tab:abs_accuracies}
\end{table}

\subsection{Single feature experiment}
\label{app:single-feat}

\begin{figure}[h]
    \centering
    \includegraphics[width=.45\linewidth]{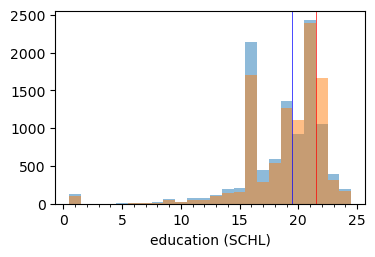}
    \includegraphics[width=.45\linewidth]{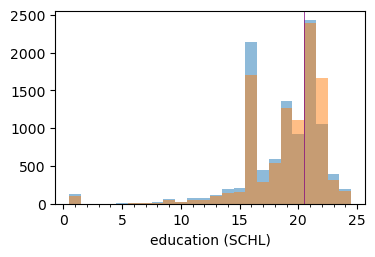}
    \caption{Histogram of the distribution of X values along the education dimension for men (blue) and women (orange/red). vertical lines denote the thresholds induced by the aware (left) and unaware (right) model.}
    \label{fig:1-feat_LR_hist}
\end{figure}

\begin{figure}[h]
    \centering
    \includegraphics[width=.49\linewidth]{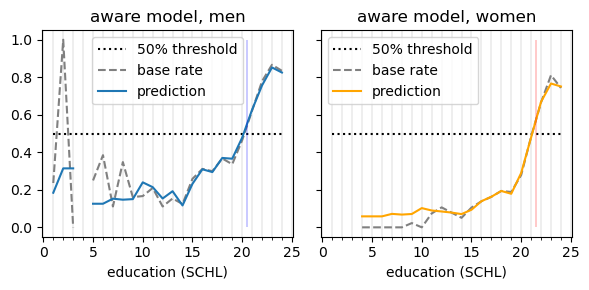}
    \includegraphics[width=.49\linewidth]{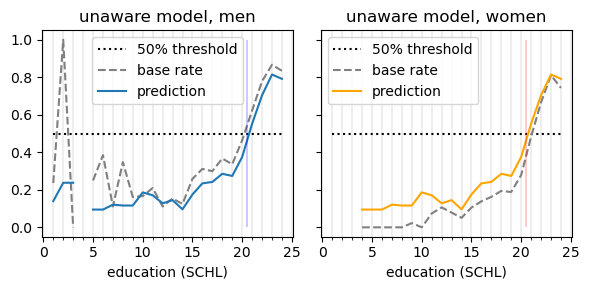}
    \caption{Predictions for aware and unaware GBM models on ACS Income NY when only using education as an input. Vertical colors indicate the 50\% cutoff, highlighting that the aware GBM model requires much higher education for women than for men.}
    \label{fig:1-feat_GBM}
\end{figure}

Here, we give more details on the single feature experiments described in Section~\ref{ss:aware_LR}.
Classification rates for the aware model are at 0.477 (men) and 0.212 (women).
Histograms of the X value distribution (Figure~\ref{fig:1-feat_LR_hist}) show that among women, there is a higher fraction of highly educated individuals, especially with a Master's degree.
This results in a higher classification rate for women (0.449) compared to men (0.389) in the unaware model, still significantly reducing the (absolute) disparate impact.
We also show model predictions for aware and unaware GBM models (Figure~\ref{fig:1-feat_GBM}).
Here, the threshold for the unaware model is the same as for LR, whereas the aware model has a higher threshold for men (21) compared with the aware LR model (same as the threshold of the unaware models).
The accuracy of the aware GBM model is slightly higher, at 70.3\%, and the classification rates are 0.389 (men) and 0.212 (women).
This still means a much higher disparate impact than for the unaware model.

\section{Proofs}
\label{app:proofs}


\subsection{Results in Section~\ref{s:calibrated}}

\begin{repproposition}{prop:brd_di}[Relation between Base Rates and Disparate Impact]\ \\
    Assume $f$ is calibrated as in (\ref{eq:calib_groups}), satisfying $\forall v \in [0,0.5): \PB(v) \geq \PA(v)$ and $\forall v \in [0.5,1] : \PA(v) \geq \PB(v)$.
    Then for $F$ defined as in (\ref{eq:thresh_class}), $\DI(F) \geq \DBR$.
\end{repproposition}
\begin{proof}
    \begin{align}
        \DI(F) &= \PA(V_a \geq 0.5) - \PB(V_b \geq 0.5) \\
        &= \int_{[0.5,1]} \PA(v) - \PB(v) \ dv \\
        &\geq \int_{[0.5,1]} (\PA(v) - \PB(v)) \cdot v \ dv \\
        & \geq \int_{[0.5,1]} (\PA(v) - \PB(v)) \cdot v \ dv + \int_{[0,0.5)} (\PA(v) - \PB(v)) \cdot v \ dv\\
        &= \int_{[0,0.5)} \PA(v) \cdot v \  dv + \int_{[0.5,1]} \PA(v)\cdot v \ dv
         - \left( \int_{[0,0.5)} \PB(v) \cdot v\  dv + \int_{[0.5,1]} \PB(v) \cdot v\ dv \right) \\
        &= \EE_{\PA}[V_a] - \EE_{\PB}[V_b] = \DBR
    \end{align}
\end{proof}
As can be seen from the proof, the assumptions in Proposition~\ref{prop:brd_di} can be relaxed to, respectively, 
\begin{equation}\label{eq:rel_ass_high}
    \int_{[0.5,1]} \PA(v) \cdot (1-v) \ dv
    \geq \int_{[0.5,1]} \PB(v) \cdot (1-v) \ dv,
\end{equation}
and
\begin{equation}\label{eq:rel_ass_low}
    \int_{[0,0.5)} \PB(v) \cdot v \ dv
    \geq \int_{[0,0.5)} \PA(v) \cdot v \ dv.
\end{equation}
The left hand side of (\ref{eq:rel_ass_low}) is equivalent to $\EE_{\PB}[V_b \giv V_b < 0.5] \cdot P(\{V_b < 0.5\})$, i.e. the mean of the predictions below the decision threshold times the probability mass below the threshold.


\subsection{Results in Section~\ref{s:ldas}}

\begin{figure}[ht]
    \centering
    \begin{tikzpicture}
        \draw[->] (0,0) -- (5.5,0) node[right] {$\lambda$};
        \draw[->] (0,0) -- (0,3.5) node[above] {$m(\lambda)$};
        
        \draw[thick] (0.4,0) rectangle (0.9,2);
        \draw[thick] (1.3,0) rectangle (1.8,3);
        \draw[thick] (2.0,0) rectangle (2.5,2.5);
        \draw[thick] (2.8,0) rectangle (3.3,1.7);
        \draw[thick] (3.55,0) rectangle (4.05,3);
        \draw[thick] (4.35,0) rectangle (4.85,1.5);
        
        \draw[thick] (0.65,0) -- (0.65,-0.1);
        \draw[thick] (1.55,0) -- (1.55,-0.1);
        \draw[thick] (3.80,0) -- (3.80,-0.1);
        \node[below] at (0.70,-0.1) {$\lambda_1$};
        \node[below] at (1.60,-0.1) {$\lambda_2$};
        \node[below] at (3.85,-0.1) {$\leps$};    
        \draw[thick] (5.1,0) -- (5.1,-0.1);
        \node[below] at (5.1,-0.1) {$0.5$};
        
        \draw[dashed] (3.65,-0.7) -- (3.65,3.4);
        
        \draw[thick, |-|] (0.05,-0.7) -- (3.45,-0.7);
        \node[below] at (1.8,-0.8) {$e(\leps)
        =\sum_{\lam < \leps} 2 \lambda \cdot m(\lambda)$};
    \end{tikzpicture}
    \caption{Illustration of $\leps$ and $e(\leps)$: $\leps$ is the highest divergence from $0.5$ s.t. switching all predictions closer than $\leps$ to $50\%$ increases the error rate by less than $\eps$.}
    \label{fig:lambda_eps2}
\end{figure}

We start with the lemma on randomised classifiers.

\begin{replemma}{lem:rand_class}\ \\
    The bound (\ref{eq:max_rash2}) can be achieved in any setting for every $\eps$ by a \emph{randomised} classifier defined as
    \begin{equation}
        F_\eps(x) = \begin{cases}
            1-B(x) & \text{ for } |p(x)-0.5| < \leps \\
            1-B(x) \text{ with probability } \frac{\eps - e(\leps)}{2 \leps \cdot m(\leps)} & \text{ for } |p(x)-0.5| = \leps \\
            B(x) & \text{ for } |p(x)-0.5| > \leps.
        \end{cases}
    \end{equation} 
\end{replemma}
\begin{proof}\ \\
    Take some $\XX$ and $P$ and $\eps$. 
    Then
    \begin{align}
        Acc(F_\eps)
        =& \sum_{\lam < \leps} m(\lam) \cdot (0.5 - \lam)
        + \sum_{\lam > \leps} m(\lam) \cdot (0.5 + \lam)\\
        &+ \left(m(\leps) \cdot \frac{\eps - e(\leps)}{2 \leps \cdot m(\leps)} \cdot (0.5 - \leps)
        + m(\leps) \cdot \left(1-\frac{\eps - e(\leps)}{2 \leps \cdot m(\leps)}\right) \cdot (0.5 + \leps)\right) \\
        =& \sum_{\lam < \leps} m(\lam) \cdot (0.5 + \lam - 2 \lam)
        + \sum_{\lam > \leps} m(\lam) \cdot (0.5 + \lam)\\
        &+ m(\leps) \cdot \frac{\eps - e(\leps)}{2 \leps \cdot m(\leps)} \cdot (0.5 + \leps - 2 \leps)
        + m(\leps) \cdot \left(1-\frac{\eps - e(\leps)}{2 \leps \cdot m(\leps)}\right) \cdot (0.5 + \leps)\\
        =& \sum_{\lam < \leps} m(\lam) \cdot (0.5 + \lam)
        + \sum_{\lam > \leps} m(\lam) \cdot (0.5 + \lam)
        - \sum_{\lam < \leps} m(\lam) \cdot 2 \lam\\
        &+ m(\leps) \cdot (0.5 + \leps)
        - m(\leps) \cdot \frac{\eps - e(\leps)}{2 \leps \cdot m(\leps)} \cdot 2\leps\\
        =& \sum_{\lam \in \Lambda} m(\lam) \cdot (0.5 + \lam)
        - e(\leps)
        - m(\leps) \cdot \frac{\eps - e(\leps)}{2 \leps \cdot m(\leps)} \cdot 2\leps\\
        =& Acc(B) - \eps
    \end{align}
    and 
    \begin{align}
        d(F_\eps,B) 
        =& \mu(\{x \in \XX : F(x) \neq B(x)\}) \\
        =& \sum_{\lam < \leps} m(\lam)
        + \frac{\eps - e(\leps)}{2 \leps \cdot m(\leps)} \cdot m(\leps) \\
        =& \frac{\eps - e(\leps)}{2 \leps} + \sum_{\lam < \leps} m(\lam).
    \end{align}
\end{proof}

This lemma helps to prove our new bound on model multiplicity.

\begin{figure}
    \centering
    \begin{tikzpicture}
        \draw[-] (0,0) -- (4.5,0) node[right] {$\XX$};
        \draw[->] (0,0) -- (0,4) node[above] {$P(Y=1|X=x)$};
        
        \node[left] at (0,2) {$0.4$};
        \node[left] at (0,2.5) {$0.5$};
        \node[left] at (0,3) {$0.6$};
        \node[left] at (0,3.5) {$0.7$};
        
        \draw[dotted] (0,2.0) -- (4.5,2.0);
        \draw[dashed] (0,2.5) -- (4.5,2.5);
        \draw[dotted] (0,3) -- (4.5,3);
        \draw[dotted] (0,3.5) -- (4.5,3.5);
        
        \draw[thick] (0.5,0) rectangle (1,2);
        \node[below] at (0.75,0) {$x_1$};
        \draw[thick] (1.5,0) rectangle (2,2);
        \node[below] at (1.75,0) {$x_2$};
        \draw[thick] (2.5,0) rectangle (3,2.75);
        \node[below] at (2.75,0) {$x_3$};
        \draw[thick] (3.5,0) rectangle (4,3.5);
        \node[below] at (3.75,0) {$x_4$};
    \end{tikzpicture}
    \hspace{1.5cm}
    \begin{tikzpicture}
        \draw[->] (0,0) -- (3.8,0) node[right] {$\lambda$}; 
        \draw[->] (0,0) -- (0,2.8) node[left] {$m(\lambda)$}; 
        
        \draw[thick] (0.5,0) rectangle (1,1); 
        \draw[thick] (1.25,0) rectangle (1.75,2);
        \draw[thick] (2.75,0) rectangle (3.25,1);
        
        \draw[thick] (0.75,0) -- (0.75,-0.1);
        \draw[thick] (1.5,0) -- (1.5,-0.1);
        \draw[thick] (3,0) -- (3,-0.1);
        \node[below] at (0.75,-0.1) {$0.05$};
        \node[below] at (1.5,-0.1) {$0.1$};
        \node[below] at (3,-0.1) {$0.2$};    
        
        \node[left] at (0,1) {$1/4$};
        \node[left] at (0,2) {$1/2$};
        \draw[thick] (0,1) -- (-0.07,1);
        \draw[thick] (0,2) -- (-0.07,2);
        
        
        \draw[|-|] (0.0,3.3) -- (2.25,3.3);
        \node[above] at (2.1,3.3) 
        {$e(0.15) = 2 \frac{0.05}{4} + 2 \frac{0.1}{2}$};
    \end{tikzpicture}
    \caption{Illustration of $e(\lambda)$ in the case of the earlier example: $e(\lambda)$ is the additional error if for all predictions closer to $50\%$ than $\lambda$, the suboptimal classification is taken. An illustration for $\lambda=0.15$ is given.}
    \label{fig:lambda_eps}
\end{figure}

\begin{figure}
    \centering
    \begin{minipage}{0.35\textwidth}
    \centering
    \begin{tabular}{c|c|c|c|c}
         & $x_1$ & $x_2$ & $x_3$ & $x_4$ \\
        \hline
        $B(x)$ & 0 & 0 & 1 & 1 \\
        \hline
        $F_1(x)$ & 0 & 0 & 0 & 1 \\
        \hline
        $F_2(x)$ & 0 & 1 & 0 & 1 \\
        \hline
        $F_3(x)$ & 1 & 1 & 1 & 1 \\
        \hline
        $F_4(x)$ & 1 & 1 & 0 & 1 \\
        \hline
    \end{tabular}
    \end{minipage}
    %
    %
    \begin{minipage}{0.5\textwidth}
        
        
        
        
        
    \begin{tikzpicture}
        \draw[->] (0,0) -- (4.4,0) node[right] {$\eps$};
        \draw[->] (0,0) -- (0,3.2) node[above] {$\max_{F \in R_\eps(B)} d(F,B)$};
    
        \node[left] at (-0.1,0.8) {$1/4$};
        \node[left] at (-0.1,1.6) {$2/4$};
        \node[left] at (-0.1,2.4) {$3/4$};
        \draw[thick] (0,0.8) -- (-0.1,0.8);
        \draw[thick] (0,1.6) -- (-0.1,1.6);
        \draw[thick] (0,2.4) -- (-0.1,2.4);
    
        \node[below] at (0.6,-0.1) {$\eps_1$};
        \node[below] at (1.8,-0.1) {$\eps_2$};
        \node[below] at (3,-0.1) {$\eps_3$};
        \draw[thick] (0.6,0) -- (0.6,-0.1);
        \draw[thick] (1.8,0) -- (1.8,-0.1);
        \draw[thick] (3,0) -- (3,-0.1);
    
        \draw[thick, black] (0,0.016) -- (2.1,2.816) node[above] {Black '22 bound};
        \draw[thick, blue] (0,-0.008) -- (0.6,0.792) -- (3,2.4) -- (4.4,2.72) node[right] {our bound};
    
        \draw[red] (0.6,0.8) circle (0.05) node[below right] {$F_1$};
        \draw[red] (1.8,1.6) circle (0.05) node[below right] {$F_2$};
        \draw[red] (2.4,1.6) circle (0.05) node[below right] {$F_3$};
        \draw[red] (3,2.4) circle (0.05) node[below right] {$F_4$};
    \end{tikzpicture}
    \end{minipage}
    \caption{Illustration of Proposition~\ref{prop:max_rash}. The table gives the classifiers derived from the calibrated predictors in Table~\ref{tab:calib_pred_example} plus an additional classifier $F_4$. We show that our bound refines the bound (\ref{eq:black_bound}) from \cite{black2022}, and is applicable under on fewer assumptions.}
    \label{fig:enter-label}
\end{figure}

\begin{repproposition}{prop:max_rash}[Upper Bound on Multiplicity]\ \\
    \begin{equation}\label{eq:max_rash2}
        \max_{F \in R_\eps(B)} d(F,B)
        \leq \frac{\eps - e(\leps)}{2 \leps} + \sum_{\lam < \leps} m(\lam)
    \end{equation}
    This bound is the lowest possible bound with only access to the cumulative distribution of $p(x)$.
\end{repproposition}

\begin{proof}\ \\
    We now show that no randomised classifier represented as $F : \XX \to [0,1]$ with $Acc(F) \leps Acc(B) - \eps$ can disagree more with $B$ on expectation than $F_\eps$. Then \emph{a forteriori}, there can be no deterministic classifier that breaks the bound.
    Now for any randomised classifier $F$ with $d(F,B) > d(F_\eps,B)$, we show $Acc(F) < Acc(F_\eps)$.
    Let the average prediction on $U(\lam)$ be given by
    \begin{equation}
         f(\lam) := \frac{1}{m(\lam)} \sum_{x \in U(\lam)} \mu(\{x\}) |F(x) - 0.5|.
    \end{equation}
    By construction of $F_\eps$,
    \begin{align}
        \forall \lam < \leps: \sum_{x \in U(\lam)} |B(x) - F_\eps(x)| \geq |B(x) - F(x)|
    \end{align}
    and
    \begin{align}
        \forall \lam > \leps: \sum_{x \in U(\lam)} |B(x) - F_\eps(x)| \leq |B(x) - F(x)|
    \end{align}
    Then $d(F,B) > d(F_\eps,B)$ implies that
    \begin{align}
        0 &\leq 
        \sum_{\lam<\leps} \sum_{x \in U(\lam)} \left( |B(x) - F_\eps(x)| - |B(x) - F(x)|\right) \cdot \mu(\{x\}) \\
        &< \sum_{\lam \geq \leps} \sum_{x \in U(\lam)} \left( |B(x) - F(x)| - |B(x) - F_\eps(x)|\right)\cdot \mu(\{x\})
    \end{align}
    But since $\forall \lam \in \Lambda$, $x \in U(\lam)$ and any randomised classifier $G$,
    \begin{equation}
        \EE_P[L(G(X), Y) \giv X=x] = \EE_P[L(B(X), Y) \giv X=x] + |B(x) - G(x)| \cdot 2 \lam
    \end{equation}
    we get
    \begin{align}
        Acc(F_\eps) - Acc(F) 
        =\ &Acc(B) - \sum_{\lam \in \Lambda} \sum_{x \in U(\lam)} |B(x) - F_\eps(x)| \cdot 2 \lam \cdot \mu(\{x\})\\
        &- Acc(B) - \sum_{\lam \in \Lambda} \sum_{x \in U(\lam)} |B(x) - F(x)| \cdot 2 \lam \cdot \mu(\{x\})\\
        =\ &\sum_{\lam \in \Lambda} \sum_{x \in U(\lam)} \left(|B(x) - F(x)| - |B(x) - F_\eps(x)|\right) \cdot 2 \lam \cdot \mu(\{x\})\\
        =\ & \sum_{\lam \geq \leps} \sum_{x \in U(\lam)} \left(|B(x) - F(x)| - |B(x) - F_\eps(x)|\right) \cdot 2 \lam \cdot \mu(\{x\}) \\
        &- \sum_{\lam<\leps} \sum_{x \in U(\lam)} \left(|B(x) - F_\eps(x)| - |B(x) - F(x)|\right) \cdot 2 \lam \cdot \mu(\{x\})\\
        \geq\ & \sum_{\lam \geq \leps} \sum_{x \in U(\lam)} \left(|B(x) - F(x)| - |B(x) - F_\eps(x)|\right) \cdot 2 \leps \cdot \mu(\{x\}) \\
        &- \sum_{\lam<\leps} \sum_{x \in U(\lam)} \left(|B(x) - F_\eps(x)| - |B(x) - F(x)|\right) \cdot 2 \leps \cdot \mu(\{x\})\\
        >\ &0.
    \end{align}
    
    We now show that the bound is tight in the sense that for any cumulative distribution (defined by $\Lambda$ and $m$), we can i) for each $\eps$ find a $\XX$, $P(X)$ and ii) for each $\XX$, $P(X)$ find an $\eps>0$ so that the bound is achievable.

    i) Take some $\Lambda$, $m$ and $\eps$.
    Then let $\XX$ be such that there is $x_0 \in U(\leps)$ with $\mu(\{x_0\}) = \frac{\eps - e(\leps)}{2 \leps}$ and let 
    \begin{equation}
        F(x) = \begin{cases}
            1-B(x) & \text{ for } |p(x)-0.5| < \leps \\
            1-B(x) & \text{ for } x = x_0 \\
            B(x) & \text{ otherwise}.
        \end{cases}
    \end{equation} 
    Then, similar to $F_\eps$ in the Lemma,
    \begin{align}
        Acc(B) - Acc(F_\eps) 
        =& \sum_{\lam \in \Lambda} \sum_{x \in U(\lam)} |B(x) - F_\eps(x)| \cdot 2 \lam \cdot \mu(\{x\})\\
        =& 2 \leps \cdot \mu(\{x_0\})
        + \sum_{\lam < \leps} 2 \lam \cdot m(\lam)\\
        =&  2 \leps \cdot \frac{\eps - e(\leps)}{2 \leps} + e(\leps)\\
        =& \eps
    \end{align}
    and
    \begin{align}
        d(F,B) 
        =& \mu(\{x \in \XX : F(x) \neq B(x)\}) \\
        =& \mu(\{x\}) + \sum_{\lam < \leps} m(\lam)  \\
        =& \frac{\eps - e(\leps)}{2 \leps} + \sum_{\lam < \leps} m(\lam).
    \end{align}
    
\end{proof}

\begin{repcorollary}{cor:black}\ \\
    Under assumption (\ref{eq:black_ass}),
    \begin{equation}\label{eq:black_cor2}
    \max_{F \in R_\eps(B)} d(F,B) \leq \frac{\eps}{2c}.
    \end{equation}
\end{repcorollary}
\begin{proof}\ \\
    From (\ref{eq:black_ass}), we get $\forall \lam \in \Lambda : \lam \geq c$ and hence
    \begin{equation}
    \frac{\eps - e(\leps)}{2 \lam} + \sum_{\lam < \leps} m(\lam)
        \leq \frac{\eps - e(\leps)}{2 c} + \sum_{\lam < \leps} m(\lam) \frac{2\lam}{2c}
        = \frac{\eps - e(\leps)}{2 c} + \frac{e(\leps)}{2 c} 
        = \frac{\eps}{2c} 
    \end{equation}
    and thus the Proposition entails
    \begin{equation}
        \max_{F \in R_\eps(B)} d(F,B)
        \leq \frac{\eps}{2c}.
    \end{equation}
\end{proof}

\begin{repproposition}{prop:two_class}[Multiplicity Bound for Non-optimal Models]\ \\
    We can bound the disagreement between two classifiers $F \in R_\eps(B)$ and $G \in R_\delta(B)$ via
    \begin{equation}\label{eq:max_rash_nonopt2}
    d(F,G) \leq \max_{H \in R_{\eps + \delta}(B)} d(H,B).
    \end{equation}
\end{repproposition}
\begin{proof}\ \\
    Fix $F$ and $G$ and let $U := \{x \in \XX : F(x) \neq G(x)\}$.
    Now define a classifier $H$ via
    \begin{equation}
        H(x) := \begin{cases}
            1-B(x) & \text{ for } x \in U \\
            B(x) & \text{ for } x \notin U.
        \end{cases}
    \end{equation}
    Then $d(B,H) = \mu(U) = d(F,G) \leq d(F,B) + d(G,B) \leq \eps + \delta$ and thus $H \in R_{\eps + \delta}(B)$.
    The result follows.
\end{proof}

\begin{repproposition}{prop:disp_imp}[Upper Bound on Disparate Impact]\ \\
    For two classifiers $F$ and $G$, the difference in their disparate impact can be bounded based on their disagreement and the size of the smaller group:
    \begin{equation}\label{eq:di_diff2}
    | \DI(F) - \DI(G) | \leq \frac{d(F,G)}{\min_{\ast \in \{\Acal,\Bcal\}} \uP{X \in \ast}}.
    \end{equation}
\end{repproposition}
\begin{proof}
    \begin{align}
        | \DI(F) - \DI(G) | 
        =\ & | \cP*{F(X)=1 \given X \in \Acal} - \cP*{F(X)=1 \given X \in \Bcal} \\
        &- \cP*{G(X)=1 \given X \in \Acal} + \cP*{G(X)=1 \given X \in \Bcal}  | \\
        =\ & | \cP*{F(X)=1 \given X \in \Acal} - \cP*{G(X)=1 \given X \in \Acal} \\
        &+ \cP*{G(X)=1 \given X \in \Bcal} - \cP*{F(X)=1 \given X \in \Bcal} |  \\
        \leq\ & \cP*{F(X) \neq G(X) \given X \in \Acal} + \cP*{F(X) \neq G(X) \given X \in \Bcal} \label{eq:di_leq_disag} \\
        \leq\ & d(F,G)\ / \min_{\ast \in \{\Acal,\Bcal\}} \uP{X \in \ast} \label{eq:di_leq_disag_min}
    \end{align}
    since
    \begin{align}
        d(F,G) 
        &= \uP{F(X) \neq G(X)} \\
        &= \cP*{F(X) \neq G(X) \given X \in \Acal} \cdot \uP{X \in \Acal} 
        + \cP*{F(X) \neq G(X) \given X \in \Bcal} \cdot \uP{X \in \Bcal}\\
        &\geq \cP*{F(X) \neq G(X) \given X \in \Acal} \cdot \min_{\ast \in \{\Acal,\Bcal\}} \uP{X \in \ast}
        + \cP*{F(X) \neq G(X) \given X \in \Bcal} \cdot \min_{\ast \in \{\Acal,\Bcal\}} \uP{X \in \ast}.
    \end{align}
\end{proof}

\begin{repproposition}{prop:unaware}[Achievable DI for Unaware Models]\ \\
    Fix $P$, an unaware model $F$, and some $\eps > 0$.
    Then there is a model $G \in R_\eps(F)$ with 
    \begin{equation}
        | \DI(F) - \DI(G) | \geq \frac{1}{2 \cdot \max_{\ast \in \{\Acal,\Bcal\}} \uP{X \in \ast}} \max_{H \in R_{\eps + \delta}(B)} d(H,B).
    \end{equation}
\end{repproposition}
\begin{proof}\ \\
    Fix an $H$ that achieves the maximum and let $U := \{x \in \XX : H(x) \neq B(x)\}$.
    Then define two classifiers $G_>$ and $G_<$ via 
    \begin{equation}
        G_>(x) := \begin{cases}
                1 & \text{ for } x \in U \cap \Acal \text{ and } F(x)=0 \\
                0 & \text{ for } x \in U \cap \Bcal \text{ and } F(x)=1 \\
                F(x) & \text{ otherwise},
            \end{cases}
    \end{equation}
    and
    \begin{equation}
        G_<(x) := \begin{cases}
                1 & \text{ for } x \in U \cap \Bcal \text{ and } F(x)=0 \\
                0 & \text{ for } x \in U \cap \Acal \text{ and } F(x)=1 \\
                F(x) & \text{ otherwise}.
            \end{cases}
    \end{equation}
    
    Then $\{x \in \XX : H(x) \neq B(x)\} = U = \{x \in \XX : F(x) \neq G_<(x)\} \cup \{x \in \XX : F(x) \neq G_>(x)\}$ and hence
    \begin{equation}
        d(F,G_\#) \geq \frac{1}{2} d(H,B)
    \end{equation}
    for one $\# \in \{<,>\}$.
    Furthermore, we get 
    \begin{equation}
        | \DI(F) - \DI(G_\#) | \geq \frac{d(F,G_\#)}{\max_{\ast \in \{\Acal,\Bcal\}} \uP{X \in \ast}}
    \end{equation}
    by the same derivation as for (\ref{eq:di_diff2}) but with equality in (\ref{eq:di_leq_disag}) due to the construction of $G_\#$ and with `$\geq$' in (\ref{eq:di_leq_disag_min}) by switching $\min$ to $\max$.
\end{proof}

\begin{repproposition}{prop:unaware_LR}[Lower Bound on DI for LR]\ \\
    Let $F : \XX \to \{0,1\}$ be an PA-aware classifier based on a LR model $f$ with coefficient $c_{G} < 0$ for the PA, i.e. the disadvantaged group gets a pre-sigmoid penalty of $c_{G}$.
    Further, let 
    \begin{equation}
        Q(c_G) := \left\{x \in \Bcal : f(x) \in [\sigma(c_{G}), 0.5]\right\}
    \end{equation}
    denote the set of points that get predictions between $\sigma(c_{G})$ and $0.5$ by $f$ \emph{and} are in Group $b$.
    Then the corresponding unaware classifier $F'$ based on setting $c_{PA} \leftarrow 0$ in $F$ has accuracy bounded by
    \begin{equation}\label{eq:unaware_LR_acc2}
        | Acc(F) - Acc(F') | \leq 2 \sigma(-c_{G}) \cdot P(X \in Q(c_G))
    \end{equation}
    and disparate impact given by
    \begin{equation}
        \DI(F') = \DI(F) - P(X \in Q(c_G)) / P(X \in \Bcal)
    \end{equation}
    if group $b$ is still disadvantaged in $F'$ (i.e. $\DI(F) \geq P(X \in Q(c_G)) / P(X \in \Bcal)$), and otherwise
    \begin{equation}\label{eq:unaware_LR_di_ineq2}
        | \DI(F) - \DI(F') | \leq P(X \in Q(c_G)) / P(X \in \Bcal).
    \end{equation}
\end{repproposition}
\begin{proof}\ \\
    $F'$ and $F$ only differ on the datapoints that are classified as $0$ by $F$ but as $1$ by $F'$.
    These are the datapoints that have prediction $<0.5$ by $f$ but $\geq 0.5$ by $f'$, i.e.
    \begin{equation}
        \{x \in \XX : \sigma^{-1}(f(x)) < 0 \wedge \sigma^{-1}(f'(x)) \geq 0\}
        = \{x \in \Bcal : - c_G \leq \sigma^{-1}(f(x)) < 0\}.
    \end{equation}
    Now this set of points is $Q(c_G)$, and the maximum change in accuracy occurs if one model is right on all points, whereas the other model is wrong on all points, \emph{and} all predictions are as extreme as possible.
    Then, assuming that one of the models is calibrated on that set, the label ratio is bounded from both sides by $\sigma(-c_{G}) \leq P(Y \giv X \in Q(c_G)) \leq \sigma(c_{G})$.
    This means the change in accuracy is bounded by $2 \cdot \sigma(c_{G}) \cdot P(X \in Q(c_G))$, giving us (\ref{eq:unaware_LR_acc2}.

    Recall that disparate impact is defined as 
    \begin{equation}
        \DI(F) = \left| P(F(X) = 1 \giv X \in \Acal) - P(F(X) = 1 \giv X \in \Bcal) \right|. 
    \end{equation}
    Now only predictions on group $b$ are changed, such that
    \begin{equation}
        P(F(X) = 1 \giv X \in \Acal) = P(F'(X) = 1 \giv X \in \Acal).
    \end{equation}
    On the other hand, we have
    \begin{equation}
        P(F'(X) = 1 \giv X \in \Bcal) = P(\{F(X) = 1 \vee - c_G \leq \sigma^{-1}(f(x)) < 0 \giv X \in \Bcal\}),
    \end{equation}
    and thus 
    \begin{equation}
         P(F'(X) = 1 \giv X \in \Bcal) =  P(F(X) = 1 \giv X \in \Bcal) +  P(X \in Q(c_G)) / P(X \in \Bcal).
    \end{equation}
    Putting things together, we get 
    \begin{equation}
        \left| P(F'(X) = 1 \giv X \in \Acal) - P(F'(X) = 1 \giv X \in \Bcal) \right|
        = \left| P(F(X) = 1 \giv X \in \Acal) - P(F(X) = 1 \giv X \in \Bcal) \right| -  P(X \in Q(c_G)) / P(X \in \Bcal)
    \end{equation}
    if the terms inside the vertical bars remain positive.
    Otherwise, we still get (\ref{eq:unaware_LR_di_ineq2}).
\end{proof}

\end{document}